%% file: main.tex
\documentclass[runningheads]{llncs}

\newcommand{\maintitle}{Improving Reasoning in Vision-Language Models via Perception Verified Self-Training}

\title{\maintitle}

\newcommand{\makesupptitle}{
    \begin{center}
        \large\bfseries \maintitle\par
        \vspace{0.5em}
        \large Supplementary Material\par
        \vspace{0.5em}
    \end{center}
}

\usepackage{eccv}

\usepackage{eccvabbrv}

\usepackage{booktabs}

\usepackage[accsupp]{axessibility}  
\usepackage{float}
\usepackage{algorithm}
\usepackage{algpseudocode}
\usepackage{amsmath}
\usepackage{amssymb}

\usepackage{tcolorbox}   
\usepackage{xcolor}      
\usepackage{graphicx}    
\usepackage{colortbl}
\definecolor{darkgreen}{RGB}{0,153,77}

\usepackage[breaklinks,colorlinks]{hyperref}

\hypersetup{
    colorlinks=true,
    allcolors=magenta   
}

\usepackage{orcidlink}

\begin{document}

\title{Improving Reasoning in Vision-Language Models via Perception Verified Self-Training} 

\titlerunning{Perception Verified Self-Training}


\author{Sourabh Sharma\inst{1}\orcidlink{0009-0008-1263-5368} \and
Sonam Gupta\inst{2}\orcidlink{0000-0003-0562-8887} \and
Sadbhawna\inst{1}\orcidlink{0000-0003-3992-6153}}

\authorrunning{S. Sharma et al.}

\institute{
Malaviya National Institute of Technology Jaipur, India
\and
IBM Software Innovation Lab, India\\
\email{sourabh125ss@gmail.com, sonam.gupta7@ibm.com, sadbhawna.cse@mnit.ac.in}}

\maketitle
\input{sections/0_abstract}    
\input{sections/1_intro}

\input{sections/2_related}    
\input{sections/3_method}

\input{sections/4_exp}

\input{sections/5_concl}
\input{sections/X_suppl}

%
%
\bibliographystyle{splncs04}
\bibliography{main}
\end{document}

%% file: sections/0_abstract.tex
\begin{abstract}

Achieving human-like reasoning in Vision-Language Models (VLMs) remains a long-standing challenge. Recent approaches leverage Chain-of-Thought (CoT) rationales generated by human annotators or proprietary models to improve reasoning, which is costly and difficult to scale. Self-training offers a promising alternative by using model's own outputs as supervision. 
However, existing methods often suffer from visual hallucinations - where rationales describe non-existent visual content, and language shortcuts - where predictions rely on textual priors rather than true visual grounding, as rationales are typically filtered only by answer correctness without verifying visual perception.
To address this limitation, we propose a perception-verified self-training framework that enforces visually grounded reasoning. First, our method employs a CoT template (caption-reasoning-conclusion) that disentangles perception from reasoning, enabling independent verification of visual understanding. To compensate for the absence of ground-truth captions, we propose \textbf{PerceptEval}, an unsupervised method that evaluates caption quality based on its alignment with visual and textual elements present in the image.
Using caption verification together with answer correctness, we partition the data into three subsets: easy (correct caption and conclusion), medium (correct caption but incorrect conclusion), and hard (incorrect caption).
Building on this partitioning, we design a \textbf{two-stage curriculum learning strategy}. In Stage 1, the model is trained on easy examples and subsequently in Stage 2, medium samples are incorporated through a \textbf{caption-guided reasoning enhancement} procedure that regenerates reasoning conditioned on verified captions. Only regenerated samples with the correct conclusions are retained. This ensures training in both stages is done exclusively on perceptually grounded reasoning, reducing hallucinations and language shortcuts.
Extensive experiments across diverse domains and models demonstrate improvements of up to 16\% over standard self-training baselines. This demonstrates that the proposed framework is a scalable and cost-effective solution to advancing multimodal reasoning without manually annotated CoT rationales.
\end{abstract}

%% file: sections/1_intro.tex
\section{Introduction}
\label{sec:intro}

\input{figures/mainfig}

Despite rapid progress in multimodal understanding \cite{stic_2024, Chen2023InternVS, li2025chemvlm, Wang2023LargescaleMP}, enhancing the reasoning capabilities of Vision-Language Models (VLMs) remains an open challenge. High-quality Chain-of-Thought (CoT) data is essential for reasoning-centric fine-tuning but is costly to obtain, particularly for multimodal tasks that require annotations grounded in both the visual and textual modalities. While proprietary models such as GPT-4V \cite{2023GPT4VisionSC} can be leveraged to generate CoT rationales, the associated cost remains prohibitive. For instance, generating only 6k detailed image descriptions may exceed \$$200$ \cite{stic_2024}. Therefore, developing scalable and cost-efficient methods for acquiring high-quality CoT data is critical for advancing VLM reasoning capabilities.

Self-training \cite{zelikman2022star, irpo, li2025care} has emerged as a promising alternative in language models, where models leverage their own generated outputs as supervision. However, extending this paradigm to VLMs introduces additional challenges. Unlike text-only models, VLMs must first correctly comprehend the visual content before reasoning about it to answer related textual queries. Existing multimodal self-training approaches, such as R3V \cite{cheng2025vision}, primarily filter training samples based on answer correctness. While effective in some cases, this strategy does not guarantee that the underlying reasoning is grounded in accurate visual perception. Consequently, the generated rationales may still contain visual hallucinations, where objects or attributes not present in the image are described, or language shortcuts, where predictions rely on textual priors instead of visual evidence. 

We argue that the key challenge in reliable self-training pipelines for VLMs lies in rectifying \textit{visual hallucinations} within their reasoning paths. To this end, we propose a perception verified self-training framework that explicitly enforces accurate perception during reasoning. Our framework employs a structured caption–reasoning-conclusion template, which first requires the model to interpret the image before generating reasoning and a final answer (conclusion). To ensure the reliability of self-generated training data, we introduce a dual-filtering mechanism that evaluates both caption quality and answer correctness.

Since ground truth for the perception component (i.e. captions) is unavailable, we propose PerceptEval, an unsupervised method for assessing the quality and correctness of generated captions. Images often contain both visual scenes and embedded textual information; therefore, PerceptEval evaluates captions through two complementary checks. First, it measures alignment between the generated caption and the image using FG-CLIP \cite{fgclip}. Second, it verifies embedded textual content using OCR-based detection \cite{paddleocr}. This combined evaluation provides a reliable signal for identifying captions that accurately reflect the image while filtering out hallucinated or incomplete descriptions.

Leveraging this perception-aware verification, we partition the generated data into subsets of varying difficulty and design a two-stage curriculum learning strategy. In the first stage, the model is trained on easy samples containing both correct captions and correct conclusions. However, relying solely on such samples limits the model’s ability to learn from cases it initially fails on. To address this limitation, we introduce caption-guided reasoning enhancement for medium-difficulty samples. For each sample with a correct caption but an incorrect conclusion, the verified caption is injected as auxiliary context when regenerating the reasoning process. Only regenerated samples that produce correct conclusions are retained for training. In the second stage, these enhanced samples are combined with the easy examples to further fine-tune the model. Importantly, training in both stages relies exclusively on samples whose reasoning is grounded in verified perception, encouraging the model to base its reasoning on visual evidence rather than exploiting language shortcuts.

Applying our framework across diverse domains such as commonsense, language science, social science and natural science leads to substantial performance gains over the baselines. Experimental results show that our approach substantially improves reasoning accuracy while reducing visual hallucinations and dependence on language shortcuts. As shown in Figure \ref{fig: mainfig}, conventional self-training pipelines often retain samples with incorrect or hallucinated rationales as long as the final answer is correct, allowing such errors to propagate during training. In contrast, our dual-verification strategy filters out these unreliable samples, ensuring that only perceptually grounded reasoning is used for learning.

To summarize, our main contributions are as follows:
(i) We introduce a perception verified self-training framework that jointly enhances visual perception and reasoning in VLMs. (ii)We propose PerceptEval, an unsupervised method for evaluating generated image captions to ensure both visual and textual aspects of the image are faithfully captured. (iii) We introduce caption-guided reasoning enhancement, a mechanism in which self-generated captions are appended to the question to guide the model towards generating image-grounded reasoning. (iv) We design a two-stage curriculum learning strategy that stabilizes iterative self-training by gradually incorporating easy and medium-difficulty samples into the training process.
\footnote{\texttt{Code: \href{https://github.com/srbhcs/perception-verified-self-training}{https://github.com/srbhcs/perception-verified-self-training}}}

%% file: figures/mainfig.tex
\begin{figure}[t]
\centering
  \includegraphics [width=\textwidth]{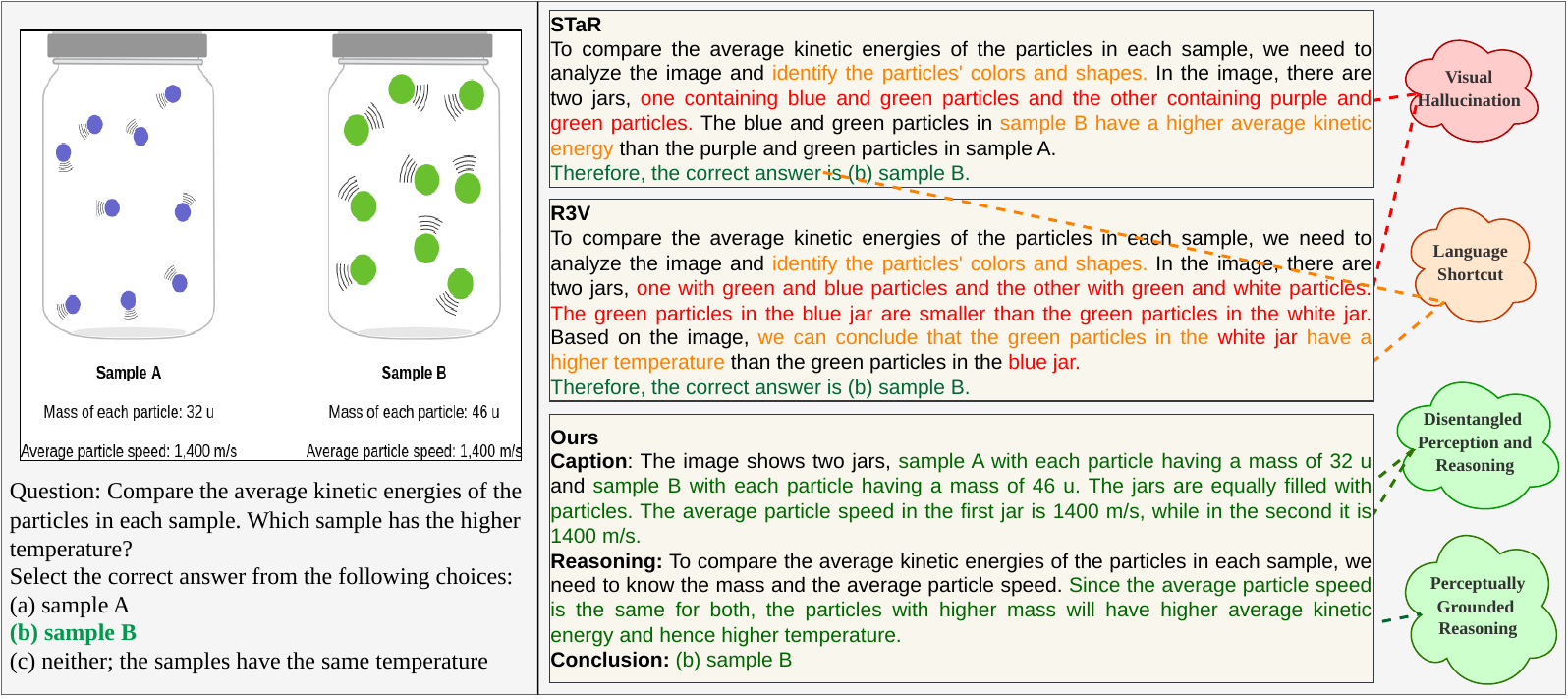}
\caption{\textbf{Comparison with STaR \cite{zelikman2022star} and R3V \cite{cheng2025vision}}. Both STaR and R3V suffer from visual hallucinations (e.g., purple particles, white jar) and language shortcuts (e.g., concluding without valid justification or inferring temperature from particle shape and color), as they filter samples solely based on final answer correctness when constructing the rationale training set. Our framework mitigates this issue through unsupervised caption verification on self-generated rationales, retaining only perceptually grounded samples for fine-tuning.}
\label{fig: mainfig}
\end{figure}

%% file: sections/2_related.tex
\section{Related Works}
\label{sec:related}

\textbf{LLMs Reasoning.} 
Early advances in LLM reasoning were driven by CoT prompting \cite{cot_llm_r1}, which enables step-by-step inference. Subsequent work showed that simple zero-shot prompts such as “Let’s think step by step” can elicit strong reasoning abilities across diverse tasks \cite{cot_prompting_llm_r2, cot_iterative_prompting_llm_r2}. Other approaches employ reward models \cite{li2023making, lu2024autopsv}, often combined with Monte Carlo Tree Search to discover high-quality reasoning paths \cite{zhang2022dino, zhang2024rest}. Reinforcement learning methods \cite{rafailov2023direct, luong2024reft} have also been explored to further improve reasoning performance.

\noindent \textbf{VLMs Reasoning.} Since human reasoning is driven by both language and visual cues, strengthening the visual reasoning capabilities of VLMs has become an essential area of exploration. Early efforts have concentrated on enhancing the reasoning capacity of VLMs through approaches like multimodal Chain-of-Thoughts(CoT) prompting strategies \cite{mitra2024compositional,dd-cot}. 
For instance, DDCoT \cite{dd-cot}improves multimodal reasoning by separating visual perception and logical reasoning, while CCoT \cite{mitra2024compositional} introduced a zero-shot prompting approach that uses scene graph representations to capture object attributes, enabling stronger compositional reasoning. Recent approaches often rely on external scaffolding like scene graphs \cite{mitra2024compositional} or bounding boxes highlighting key region in images \cite{shao2024visual}. More recently, reinforcement learning (RL) has emerged as an effective paradigm for enhancing reasoning in VLMs. Vision-R1 \cite{VisionR1} combines supervised cold-start initialization with GRPO to improve multimodal reasoning. Visionary-R1 \cite{VisionaryR1} proposes a caption-reason-answer optimization strategy to encourage explicit visual interpretation and mitigate shortcut learning under pure RL training. Vision-SR1 \cite{VisionSR1} introduces a self-rewarding multi-reward RL framework that separately optimizes visual perception and language reasoning to reduce visual hallucinations and language shortcuts. These methods demonstrate the effectiveness of RL in improving visual reasoning across challenging multimodal tasks. However, they generally require computationally expensive policy optimization, rollout sampling, and carefully designed reward mechanisms. In contrast, our work adopts a lightweight perception-verified self-training framework based solely on supervised fine-tuning, improving visual grounding without RL.


\noindent \textbf{Self-training for Reasoning.} 
Self-training improves models by learning from their own generated data and has shown strong success for LLMs \cite{zelikman2022star, irpo, zhang2024rest}. STaR \cite{zelikman2022star} introduced a self-taught reasoning loop where models iteratively generate and refine their own Chain-of-Thought rationales using ground-truth answers as hints. While effective for text-only tasks, extending self-training to visual reasoning remains challenging due to visual hallucinations in generated reasoning paths.
\input{figures/arch}
Recent work has begun exploring self-training for VLMs \cite{cheng2025vision, zohar2024video, stic_2024}. R3V \cite{cheng2025vision} improves reasoning through iterative reflection on generated rationales. Video-STaR \cite{zohar2024video} applies self-training to video understanding using labeled video datasets, while STIC \cite{stic_2024} constructs preference data from self-generated responses to enhance image understanding. Despite these advances, current approaches typically rely on task-specific supervision or implicit textual priors, and often do not explicitly verify whether intermediate perceptions are grounded in the visual input that leads to visual hallucinations and weak visual grounding. To address this limitation, we propose a perception-verified self-training framework that encourages visually grounded reasoning through disentangled perception–reasoning prompts and dual-filtering mechanisms.


%% file: figures/arch.tex
\begin{figure*}
    \centering
    \includegraphics[width=\linewidth]{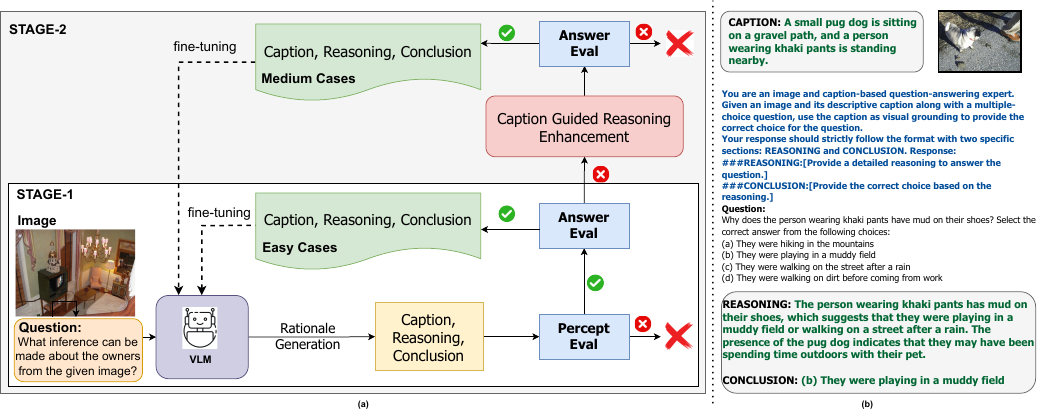}
    \caption{\textbf{(a) Overview of our proposed framework.} Dashed arrows indicate fine-tuning steps. Stage-1 fine-tuning uses only easy cases, while Stage-2 incorporates both easy and medium cases. AnswerEval verifies whether the generated conclusion matches the ground-truth answer, and PerceptEval assesses caption quality in the absence of ground-truth captions. \textbf{(b) Example of a medium case where the model initially produces incorrect reasoning.} By applying caption-guided reasoning enhancement enables the model to correct its reasoning and produce the correct conclusion.}
    \label{fig:arch}
\end{figure*}

%% file: sections/3_method.tex
\section{Method}

Let $M$ be an instruction-tuned Vision–Language Model (VLM), and let 
$\mathcal{D}=\{(I_i, q_i, a_i)\}_{i=1}^{N}$ denote a dataset of $N$ image–question–answer triplets, where $I_i$, $q_i$, and $a_i$ represent the $i^{th}$ image, question, and ground-truth answer, respectively. 
Self-training frameworks augment this dataset by a rationale $\hat{r}_i$ for each sample. Only those rationales are retained whose conclusion matches the ground truth answer. This process transforms $\mathcal{D}$ into a visual reasoning dataset $R = \{(I_i, q_i, \hat{d}_i, \hat{r}_i, a_i)\}_{i=1}^{N'}$. However, as illustrated in Figure~\ref{fig: mainfig}, filtering samples solely based on answer correctness allows contaminated rationales containing visual hallucinations to enter the training set, leading to suboptimal reasoning and degraded model performance. To address this limitation, we propose a perception-verified self-training framework that tackles two key challenges: (1) verifying the correctness of the model’s visual perception, and (2) leveraging correct perception to ground reasoning.

To address the first challenge, we adopt a structured \textsc{[Cap–Reas–Concl]} template that explicitly separates perception from reasoning, making the perception component independently verifiable. Specifically, the model is prompted to produce a tuple \( (\hat{d}_i, \hat{r}_i, \hat{a}_i) \), consisting of an image description \( \hat{d}_i \), reasoning \( \hat{r}_i \), and the predicted answer \( \hat{a}_i \). The exact prompt used to elicit this structured output is provided in the supplementary material (SM). Since ground-truth captions are unavailable, we introduce the PerceptEval module to assess caption fidelity. Section~\ref{sec: percepteval} details the PerceptEval pipeline for validating caption quality.

To address the second challenge, we combine caption verification with answer correctness to partition the generated samples into three categories: \textit{easy} (correct caption and correct conclusion), \textit{medium} (correct caption but incorrect conclusion), and \textit{hard} (incorrect caption). Based on this partitioning, we design a two-stage curriculum learning strategy and introduce a \textit{caption-guided reasoning enhancement} mechanism that grounds reasoning in verified captions for medium-difficulty samples, producing more faithful and accurate rationales. Section~\ref{sec: curriculum} details this process.

\subsection{Caption Validation via PerceptEval}
\label{sec: percepteval}

We propose \textbf{PerceptEval}, an unsupervised perception validation module that evaluates the fidelity of generated captions. To ensure robust validation across both text-heavy images (e.g., scanned documents) and visually rich images, PerceptEval relies on two complementary feedback signals. The first signal measures the similarity between the model-generated caption and OCR-extracted text, ensuring that captions faithfully capture textual content present in the image. The second signal computes the \texttt{FG-CLIP} \cite{fgclip} similarity between the caption and the image, providing a holistic measure of visual consistency. Together, these signals enable robust filtering of high-quality captions prior to self-training.
\input{figures/percept_eval}

\input{algorithms/algomain_1}

\noindent\textbf{OCR–Based Text Agreement:} 
Many images in multimodal VQA tasks contain textual information that is essential for answering the question. We employ \texttt{PaddleOCR} \cite{paddleocr} to extract all visible text from the image and construct an auxiliary caption $d_i^{\text{ocr}}$ of the form:
\textit{"The image has text written as: \{OCR text\}."}
We then compute the semantic similarity $s_i^{\text{ocr}}$ between the OCR-derived caption $d_i^{\text{ocr}}$ and the model-generated caption $\hat{d}_i$ using \texttt{Sentence-Transformers} \cite{sentformer} embeddings with cosine similarity. A high similarity score indicates that the generated caption accurately reflects the textual information embedded in the image.

\noindent\textbf{Visual Agreement with \texttt{FG-CLIP}:} 
To evaluate visual alignment, we compute the image–caption similarity score $s_i^{\text{vis}}$ using \texttt{FG-CLIP} \cite{fgclip}. A high similarity score indicates that the caption $\hat{d}_i$ correctly describes objects, scenes, or entities visible in the image.

\noindent\textbf{Domain-Aware Joint Evaluation.}
Since images vary in the proportion of textual versus visual content, applying fixed similarity thresholds across all samples may lead to suboptimal filtering. To address this, we compute the text area coverage ratio:

\[
r_t = \frac{\text{Total OCR box area}}{\text{Image area}},
\]

\noindent where text bounding boxes are obtained via \texttt{PaddleOCR}~\cite{paddleocr}. Based on $r_t$, our framework adaptively adjusts similarity thresholds for the OCR and \texttt{FG-CLIP} scores. Additional details are presented in SM.

Samples $(\hat{d}_i, \hat{r}_i, \hat{a}_i)$ that do not satisfy these content-aware thresholds are filtered out, resulting in a subset of samples with reliable perceptual alignment. An illustration of the PerceptEval pipeline is shown in Figure~\ref{fig: percepteval}.

\subsection{Self-Training with Two-Stage Curriculum Learning}
\label{sec: curriculum}

After validating captions, we assess the correctness of the model’s reasoning by comparing the predicted conclusion $\hat{a}_i$ with the ground-truth answer $a_i$. Based on caption fidelity and answer correctness, we partition the samples into three subsets: \textit{easy} $\mathcal{D}_{\text{easy}}$ (correct caption and correct conclusion), \textit{medium} $\mathcal{D}_{\text{medium}}$ (correct caption but incorrect conclusion), and \textit{hard} $\mathcal{D}_{\text{hard}}$ (incorrect caption).

\noindent \textbf{Stage 1: Training on Easy Samples.}
In the first stage, we fine-tune the model using the high-quality easy subset $\mathcal{D}_{\text{easy}}$. After each training round, the model regenerates rationales for the dataset, gradually producing higher-quality supervision. To maintain training stability and avoid overfitting, fine-tuning is always performed on the original model $M$ rather than the previous checkpoint. Because the supervision follows the \textsc{[Cap–Reas–Concl]} structure, finetuning improves both visual perception and reasoning while ensuring that the reasoning remains grounded in the image. We repeat this self-training loop (Algorithm~\ref{alg:stage_1}) until validation performance plateaus.

\input{algorithms/algomain_2}

\noindent \textbf{Stage 2: Incorporating Medium Samples.}
Training solely on easy samples eventually limits the model’s ability to solve more challenging problems. To expand the supervision signal, the second stage incorporates the \textit{medium} subset, i.e., samples with correct captions but incorrect reasoning. These cases indicate that the model has correctly perceived the visual content but fails to effectively utilize this perception during reasoning. We convert such samples into training data through a caption-guided reasoning enhancement procedure.

\noindent \textbf{Caption-Guided Reasoning Enhancement.}
For each medium sample $(\hat{d}_i,\allowbreak\ \hat{r}_i,\allowbreak\ \hat{a}_i)$, the verified caption $\hat{d}_i$ is injected as additional context to guide reasoning. Specifically, we employ a caption-embedded prompting strategy (Figure~\ref{fig:arch}(b)) that conditions the model to generate responses in the \textsc{[Reas–Concl]} format grounded in the caption. The generated responses are then validated using answer correctness. If the predicted conclusion matches the ground-truth answer, the newly generated reasoning is combined with the verified caption to construct refined $(\hat{d}_i, \Tilde{r}_i, a_i)$ training instances. These validated medium samples $\mathcal{D}_{\text{med}}$ are then merged with the easy subset to form the training set for the next iteration, i.e., $\mathcal{D}_{\text{easy}} \cup \mathcal{D}_{\text{med}}$.  We repeat this stage 2 self-training loop (Algorithm~\ref{alg:stage_2}) until validation performance plateaus. Figure~\ref{fig:arch}(a) illustrates the overall flow of the proposed framework.

%% file: figures/percept_eval.tex
\begin{figure}[t]
\centering
  \includegraphics [width=.9\textwidth]{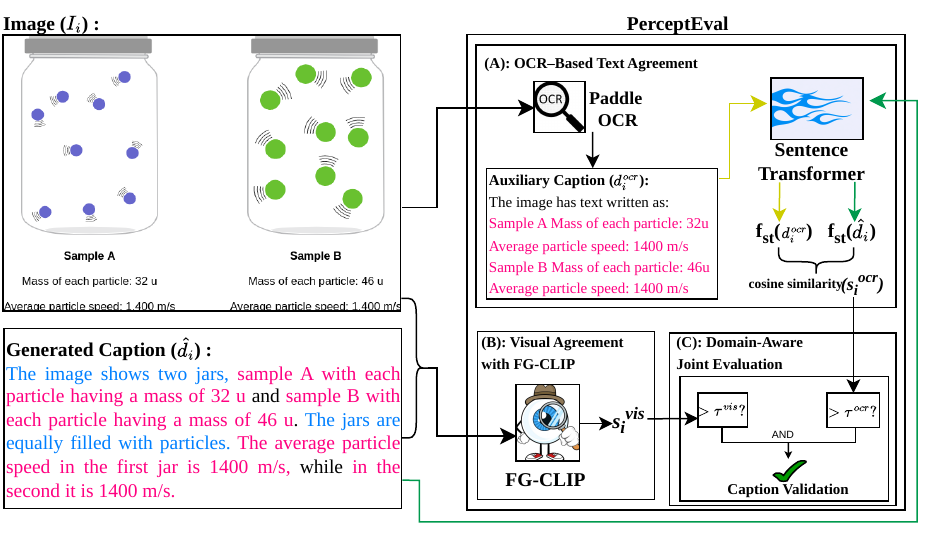}
\caption{\textbf{Illustration of PerceptEval.} OCR agreement is computed as the cosine similarity between the sentence-transformer embeddings \cite{sentformer} of the generated caption and an auxiliary caption obtained from PaddleOCR \cite{paddleocr}. Visual similarity is measured using FG-CLIP \cite{fgclip} between the image and the generated caption. FG-CLIP \cite{fgclip} focuses on visual element descriptions in the caption (blue), while OCR alignment ensures the inclusion of extracted text (pink). A caption is accepted only if both scores exceed their respective domain-adaptive thresholds.}
 
\label{fig: percepteval}
\end{figure}

%% file: algorithms/algomain_1.tex
\begin{algorithm}
\caption{\textsc{Stage 1: Self-Training with Easy Cases}}
\label{alg:stage_1}
\begin{algorithmic}[1]
\Require Pretrained VLM $M$; Dataset $\mathcal{D}=\{(I_i, q_i, a_i)\}_{i=1}^{N}$
\State Initialize $M_0 \gets M$, $t \gets 0$
\Repeat
    \State $t \gets t + 1$
    \ForAll{$(I_i, q_i, a_i) \in \mathcal{D}$}
        \State $(\hat{d}_i, \hat{r}_i, \hat{a}_i) \gets M_{t-1}[\texttt{CRC-Prompt}, I_i, q_i]$
        \State $\text{valid}^{\text{cap}}_i \gets \textsc{PerceptEval}(I_i, \hat{d}_i)$ 
        \State $\text{valid}^{\text{ans}}_i \gets (\hat{a}_i = a_i)$ 
        \If{$\text{valid}^{\text{cap}}_i \land \text{valid}^{\text{ans}}_i$}
            \State $\tilde{x}_i \gets (\texttt{CRC-Prompt}, I_i, q_i, \hat{d}_i, \hat{r}_i, a_i)$
            \State Add $\tilde{x}_i$ to $\mathcal{D}_{\text{easy}}$
        \EndIf
    \EndFor
    \State $M_t \gets \texttt{train}(M, \mathcal{D}_{\text{easy}})$
\Until{converged}
\State $M_{\text{stage1}} \gets M_t$
\State \Return $M_t$
\end{algorithmic}
\end{algorithm}

%% file: algorithms/algomain_2.tex
\begin{algorithm}
\caption{\textsc{Stage 2: Mixed Self-Training with Medium Cases}}
\label{alg:stage_2}
\begin{algorithmic}[1]
\Require Pretrained VLM $M$; Dataset $\mathcal{D}=\{(I_i, q_i, a_i)\}_{i=1}^{N}$
\State Initialize $M_0 \gets M_{\text{stage1}}, t \gets 0$
\Repeat
    \State $t \gets t + 1$
    \ForAll{$(I_i, q_i, a_i) \in \mathcal{D}$}
        \State $(\hat{d}_i, \hat{r}_i, \hat{a}_i) \gets M_{t-1}[\texttt{CRC-Prompt}, I_i, q_i]$
        \State $\text{valid}^{\text{cap}}_i \gets \textsc{PerceptEval}(I_i, \hat{d}_i)$
        \State $\text{valid}^{\text{ans}}_i \gets (\hat{a}_i = a_i)$
        \If{$\text{valid}^{\text{cap}}_i \land \text{valid}^{\text{ans}}_i$}
            \State $\tilde{x}_i \gets (\texttt{CRC-Prompt}, I_i, q_i, \hat{d}_i, \hat{r}_i, a_i)$
            \State Add $\tilde{x}_i$ to $\mathcal{D}_{\text{easy}}$
        \ElsIf{$\text{valid}^{\text{cap}}_i \land \lnot \text{valid}^{\text{ans}}_i$}
            \State $(\tilde{r}_i, \tilde{c}_i) \gets M_{t-1}[\texttt{CapEmb-Prompt}, I_i, q_i]$
            \If{$\tilde{c}_i = a_i$}
                \State $\tilde{y}_i \gets (\texttt{CRC-Prompt}, I_i, q_i, \hat{d}_i, \tilde{r}_i, a_i)$
                \State Add $\tilde{y}_i$ to $\mathcal{D}_{\text{med}}$
            \EndIf
        \EndIf
    \EndFor
    \State $M_t \gets \texttt{train}(M, \mathcal{D}_{\text{easy}} \cup \mathcal{D}_{\text{med}})$
\Until{converged}
\State \Return $M_t$
\end{algorithmic}
\end{algorithm}

%% file: sections/4_exp.tex
\section{Experiments}
\label{sec:experiments}

This section first presents our experimental setup, including the base models, datasets, and baseline methods used for comparison, along with key implementation details in Subsection \ref{sec:setup}.
It then discusses the quantitative results and provides a detailed qualitative analysis in Subsection \ref{sec:main}. Subsection \ref{sec:additional_analysis} presents additional insights and quantitative analyses. Subsection \ref{sec: ablation} presents the ablation studies and provides a detailed discussion of the results.

\subsection{Experiment Setup}
\label{sec:setup}

\noindent \textbf{Models.} To evaluate the effectiveness of our approach, we conduct experiments across four knowledge domains with the \textit{LLaVA-v1.5-7B} \cite{liu2023improvedllava} model. To assess generalization across different vision-language models, we also evaluate our method with \textit{Qwen2-VL-7B-Instruct} \cite{Qwen2VL}. Finally, to examine scalability to larger models, we also evaluate our method with \textit{LLaVA-v1.5-13B} \cite{liu2023improvedllava} model.

\noindent \textbf{Dataset.} We evaluate our method across four domains, namely commonsense, natural science, language science, and social science, using samples from the M3CoT dataset \cite{chen-etal-2024-m3cot}, which provides multi-domain, multiple-choice visual question-answering. Each domain includes train, validation and test splits. Further, for Out-of-Distribution evaluation, we have evaluated on the science domain of MMMU \cite{yue2024mmmu} validation split.


\noindent \textbf{Baselines.} To comprehensively evaluate the reasoning capabilities of our method, we compare it against a diverse set of baselines. For zero-shot evaluation, we measure the base model’s performance under various prompting strategies including, direct answer prompting, Chain-of-Thought (CoT) prompting (lets think step by step) \cite{letsthinkstepbystep}, Desp-CoT \cite{wu2023rolechainofthoughtcomplexvisionlanguage}, CCoT \cite{mitra2024compositionalchainofthoughtpromptinglarge} and CRC prompt template. We also include a direct SFT baseline, where the model is fine-tuned on (image, question, answer) tuples using direct prompting, instructing it to predict the answer directly rather than generating an intermediate rationale. Finally, we compare our method with two state-of-the-art self-training methods, STaR \cite{zelikman2022star} and R3V \cite{cheng2025vision}, which represent the leading approaches for rationale-based self-improvement, providing a strong baseline to assess the effectiveness of our self-training framework.

\noindent \textbf{Implementation Details.} All the models are trained with an effective batch size of 16, LR 3e-5 for 2 epochs using AdamW; LoRA settings were (rank 64, alpha 16) for \textit{Qwen-2-VL} \cite{Qwen2VL} and (rank 128, alpha 256) for \textit{LLaVA-1.5} \cite{liu2023improvedllava}, with 0.05 dropout. All experiments were conducted on a single NVIDIA A6000 GPU with 48 GB of VRAM. More experimental details are presented in the SM.

\subsection{Main Results}
\label{sec:main}

\input{tables/main_llava}
\input{tables/main_qwen}
\input{tables/joint_table}
\input{figures/qual_1}
\input{figures/joint}

\textbf{Quantitative Analysis.}
We compare our method against a diverse set of baselines using test-set accuracy, with results for LLaVA summarized in Table~\ref{tab:llava_methods}. In zero-shot setting, VLMs exhibit clear limitations in multimodal reasoning. Notably, CoT prompting performs worse than direct answer prompting (Direct VQA). This suggests that VLMs tend to rely on shortcut patterns rather than genuine reasoning. When prompted to reason step-by-step, models such as LLaVA often generate incorrect rationales leading to wrong answers. Whereas, direct prompting avoids explicit reasoning and therefore achieves slightly higher accuracy. Fine-tuning the model with a direct SFT baseline on (image, question, answer) tuples yields only modest improvements over zero-shot methods.

Existing self-training approaches such as STaR \cite{zelikman2022star} and R3V \cite{cheng2025vision} also show limited gains. STaR uses a hint-based augmentation strategy that exposes the model to the correct answer, which can encourage shortcut learning and introduce noisy rationales. R3V selects samples solely based on final answer correctness, causing perception and reasoning to become entangled while optimization focuses only on answer prediction, leading to early saturation.

In contrast, our method achieves consistent gains across all domains. By explicitly separating perception and reasoning in the structured prompt and filtering samples based on the correctness of both signals, our framework ensures higher-quality rationales for training. This enables the model to optimize both components during fine-tuning, resulting in more reliable multimodal reasoning and improved overall performance.

To further evaluate the generalizability of our approach to stronger, modern VLMs, we applied the proposed framework to \textit{Qwen-2-VL-7B} (Table \ref{tab:qwen_methods}). Compared to LLaVA, Qwen exhibits superior reasoning capabilities, as reflected in Direct VQA performance. However, shortcut learning persists: performance drops significantly when the model is prompted to generate reasoning before answering. Even in this setting, our method outperforms STaR and R3V, demonstrating the robustness and effectiveness of our framework in improving reasoning quality and overall performance across different model families.

\noindent\textbf{Qualitative Analysis.}
We visualize the rationales generated on the test sets of Social Science (Figure~\ref{qual4} (a)) and Commonsense (Figure~\ref{qual4} (b)) domains, qualitatively comparing our method with existing self-training approaches, STaR and R3V to gain deeper insights. In Figure~\ref{qual4} (a), we observe that due to the absence of perception verification during self-training, STaR \cite{zelikman2022star} and R3V \cite{cheng2025vision} learn to produce rationales with either no visual details or hallucinated descriptions. In contrast, our method generates rationales that are both visually detailed and logically coherent.
Figure~\ref{qual4}(b) shows that STaR \cite{zelikman2022star} and R3V \cite{cheng2025vision} misinterpret the scene, associating the camera recording with a job interview instead of attending to the visual details, thereby exhibiting language shortcuts. In contrast, our method correctly identifies key objects, such as the suitcase, grounding the reasoning and leading to the correct answer. More results are presented in SM.

\subsection{Additional Analyses}
\label{sec:additional_analysis}
\noindent \textbf{Out-of-Distribution (OOD) Evaluation.}
To examine whether the improved perception and grounded multimodal reasoning capabilities learned through our approach generalize to OOD settings, we evaluate the baselines on the science validation split of the MMMU dataset \cite{yue2024mmmu}. Zero-shot methods are evaluated on the base model \textit{LLaVA-v1.5-7B} \cite{liu2023improvedllava}. 
For self-training methods, the base model is first trained on the natural science domain of M3CoT \cite{chen-etal-2024-m3cot} via the respective self-training framework. As shown in Table~\ref{tab:mmmu_ns}, self-training improves the model’s general reasoning ability compared to the zero-shot baseline. Furthermore, our proposed framework achieves the best performance compared to other self-training methods. By incorporating only high-quality rationales grounded in verified perceptual evidence, our approach enhances both perceptual grounding and multimodal reasoning, resulting in stronger generalization to OOD tasks.

\noindent \textbf{Scalability.} We evaluate the scalability of our framework to larger model sizes by conducting experiments with the \textit{LLaVA-v1.5-13B} \cite{liu2023improvedllava} model on the Language Science and Commonsense domains. The results are reported in Table~\ref{tab:generalize_llava_methods}. Compared to its smaller counterpart, the larger model shows greater difficulty in strictly adhering to the CRC template under zero-shot prompting. Nevertheless, the overall trend remains consistent, indicating that our framework continues to provide improvements even at larger model scales.

\input{tables/hallucination}
\noindent \textbf{Visual Hallucination Evaluation.}
To evaluate the effectiveness of our framework in mitigating visual hallucinations, we assess \textit{LLaVA-v1.5-7B} \cite{liu2023improvedllava} model trained on the Language Science domain of M3CoT~\cite{chen-etal-2024-m3cot} under different settings on the OCR split of HallusionBench~\cite{Guan_2024_CVPR}, which serves as an Out-of-Distribution benchmark within the same domain. As shown in Table~\ref{tab:hallusionbench_ocr}, our method consistently outperforms the baseline and existing self-training approaches. These results suggest that perception verification enables the model to generate reasoning chains that are more faithfully grounded in the visual input, thereby reducing visual hallucinations.

\noindent \textbf{Token Efficiency.} We compute the average output token length across the test splits of the M3CoT \cite{chen-etal-2024-m3cot} domains using the \textit{LLaVA-v1.5-7B} \cite{liu2023improvedllava} model. STaR \cite{zelikman2022star} generates shorter and often less detailed rationales, resulting in brief outputs (31.6–56.1 tokens). In contrast, R3V produces substantially longer responses (65.6–126.0 tokens). Our method yields moderate length (69.2–93.9 tokens) outputs, closely aligning with the base model (63.6–105.0 tokens), thereby balancing explicit reasoning with token efficiency. Detailed results are provided in SM.

\noindent \textbf{Subjective Analysis.} To evaluate reasoning quality subjectively, we follow the STaR \cite{zelikman2022star} setup. For each M3CoT \cite{chen-etal-2024-m3cot} domain, we sample 50 instances where all methods produce the correct answer. Each instance includes four anonymized rationales, which are ranked by three experts based on reasoning quality (1=best, 4=worst). Rankings are aggregated using Borda score and visualized as a heat map (Figure ~\ref{fig:heatmap}). Our method consistently attains the highest reasoning quality, indicated by darkest cells. Annotation details are presented in SM. 

\subsection{Ablation Studies}
\label{sec: ablation}
\input{tables/joint_ablation}
We conduct three ablation studies to analyze the contributions of different components in our framework.  
\textbf{First}, Table \ref{tab:ablation_cot_caption} shows how each component contributes to our framework’s performance. Using answer-based filtering alone performs poorly, as it provides no feedback to improve captions during optimization. Incorporating \textsc{PerceptEval} boosts results, but performance remains limited because the model is trained only on easy samples. Simply adding medium-difficulty samples from the start leads to early saturation, with the model struggling to learn easy and medium cases simultaneously. By contrast, our two-stage curriculum learning strategy introduces medium-difficulty samples gradually, allowing the model to build on simpler cases first and achieve higher performance gains.

\textbf{Second}, Table \ref{tab:ablation_ocr_fgclip} highlights the contributions of the two components of \textsc{PerceptEval}: OCR similarity and FG-CLIP similarity. For text-heavy language-science tasks, FG-CLIP alone misses fine-grained OCR details. Conversely, using only OCR similarity can also fail when captions misinterpret OCR text as objects in the image. Combining FG-CLIP and OCR similarity mitigates these issues, as illustrated in Figure \ref{textfig}. For commonsense tasks, where images are visually dominated, OCR agreement alone cannot capture the relevant visual context. Balancing both components is therefore crucial for achieving strong performance.

\textbf{Third}, Figure \ref{fig:validation_accuracy} illustrates the effectiveness of our two-stage curriculum learning. Introducing medium-difficulty samples from the first iteration, i.e., without curriculum learning, confuses the model, often resulting in early saturation or failure. In contrast, our approach initially trains the model on easy samples until performance saturates, and only then introduces medium-difficulty samples. As shown in the figure, the validation accuracy improves sharply when medium samples are introduced at iteration 8. This staged approach allows the model to master fundamentals before tackling harder examples, yielding substantial performance gains in subsequent iterations.
\input{figures/graph}

%% file: tables/main_llava.tex
\begin{table}[t]
\centering
\caption{Comparison of VQA performance (\%) across reasoning methods and domains on \textit{LLaVA-v1.5-7B} \cite{liu2023improvedllava}.}
\setlength{\tabcolsep}{4.2pt}
\renewcommand{\arraystretch}{1.1}
\begin{scriptsize}
\begin{tabular}{lcccc}
\toprule
\textbf{Method} & \textbf{Lang. Sci.} & \textbf{Common.} & \textbf{Social Sci.} & \textbf{Natural Sci.} \\
\midrule
\rowcolor{gray!10}
\multicolumn{5}{l}{\textbf{Zero-Shot Methods}} \\
\texttt{Direct VQA} & 45.5 & 57.58 & 29.62 & 36.4 \\
\texttt{CoT\cite{letsthinkstepbystep}} & 38.86 & 59.34 & 25.48 & 33.59 \\
\texttt{Desp-CoT\cite{wu2023rolechainofthoughtcomplexvisionlanguage}} & 34.12 & 54.73 & 25.32 & 32.18 \\
\texttt{CCoT\cite{mitra2024compositionalchainofthoughtpromptinglarge}} & 26.54 & 54.08 & 28.66 & 35.5 \\
\texttt{[CAP-REAS-CONCL]} & 40.28 & 53.4 & 27.55 & 33.2 \\
\midrule
\rowcolor{gray!10}
\multicolumn{5}{l}{\textbf{Direct SFT}} \\
\texttt{VQA Finetune} & 46.45 & 60.22 & 34.24 & 46.1 \\
\midrule
\rowcolor{gray!10}
\multicolumn{5}{l}{\textbf{Self-Training Methods}} \\
\texttt{STaR\cite{zelikman2022star}} & 48.82 & 64.98 & 41.88 & 53.9 \\
\texttt{R3V\cite{cheng2025vision}} & 50.24 & 65.49 & 42.2 & 51.34 \\
\texttt{\textbf{Ours}} & \textbf{64.93} & \textbf{72.74} & \textbf{46.56} & \textbf{58.24} \\
\bottomrule
\end{tabular}
\end{scriptsize}
\label{tab:llava_methods}
\end{table}

%% file: tables/main_qwen.tex
\begin{table}[t]
\centering
\caption{Comparison of VQA performance (\%) across reasoning methods and domains on \textit{Qwen-2-VL-Instruct} \cite{Qwen2VL}.}
\setlength{\tabcolsep}{4.2pt}
\renewcommand{\arraystretch}{1.1}
\begin{scriptsize}
\begin{tabular}{lcccc}
\toprule
\textbf{Method} & \textbf{Lang. Sci.} & \textbf{Common.} & \textbf{Social Sci.} & \textbf{Natural Sci.} \\
\midrule
\rowcolor{gray!10}
\multicolumn{5}{l}{\textbf{Zero-Shot Methods}} \\
\texttt{Direct VQA} & 68.72 & 77.8 & 38.85 & 54.15 \\
\texttt{[CAP-REAS-CONCL]} & 63.03 & 54.72 & 37.58 & 47.13 \\
\midrule
\rowcolor{gray!10}
\multicolumn{5}{l}{\textbf{Direct SFT}} \\
\texttt{VQA Finetune} & 76.78 & 80.22 & 45.22 & 65.39 \\
\midrule
\rowcolor{gray!10}
\multicolumn{5}{l}{\textbf{Self-Training Methods}} \\
\texttt{STaR\cite{zelikman2022star}} & 75.83 & 78.02 & 46.5 & 66.67 \\
\texttt{R3V\cite{cheng2025vision}} & 76.78 & 78.68 & 44.9 & 63.86 \\
\texttt{\textbf{Ours}} & \textbf{78.67} & \textbf{82.42} & \textbf{49.04} & \textbf{68.45} \\
\bottomrule
\end{tabular}
\end{scriptsize}
\label{tab:qwen_methods}
\end{table}

%% file: tables/joint_table.tex
\begin{table}[t]
\centering

\begin{minipage}{0.48\linewidth}
\centering
\caption{Evaluation on the MMMU Science validation subset.}
\setlength{\tabcolsep}{8pt}
\renewcommand{\arraystretch}{1.3}
\begin{scriptsize}
\begin{tabular}{lc}
\toprule
\textbf{Method} & \textbf{Accuracy} \\
\midrule
\rowcolor{gray!10}
\multicolumn{2}{l}{\textbf{Zero-Shot Methods}} \\
\texttt{Zero-Shot VQA} & 21.62 \\
\texttt{CoT\cite{letsthinkstepbystep}} & 27.02 \\
\texttt{[CAP-REAS-CONCL]} & 28.37 \\
\midrule
\rowcolor{gray!10}
\multicolumn{2}{l}{\textbf{Self-Training Methods}} \\
\texttt{STaR\cite{zelikman2022star}} & 32.43 \\
\texttt{R3V\cite{cheng2025vision}} & 33.78 \\
\texttt{\textbf{Ours}} & \textbf{36.48} \\
\bottomrule
\end{tabular}
\end{scriptsize}
\label{tab:mmmu_ns}
\end{minipage}
\hfill
\begin{minipage}{0.48\linewidth}
\centering
\caption{Comparison of VQA performance (\%) across reasoning methods on \textit{LLaVA-v1.5-13B} \cite{liu2023improvedllava}.}
\setlength{\tabcolsep}{4.2pt}
\renewcommand{\arraystretch}{1}
\begin{scriptsize}
\begin{tabular}{lcc}
\toprule
\textbf{Method} & \textbf{Lang. Sci.} & \textbf{Common.} \\
\midrule
\rowcolor{gray!10}
\multicolumn{3}{l}{\textbf{Zero-Shot Methods}} \\
\texttt{Direct VQA} & 45.97 & 66.15 \\
\texttt{[CAP-REAS-CONCL]} & 32.7 & 41.76 \\
\midrule
\rowcolor{gray!10}
\multicolumn{3}{l}{\textbf{Direct SFT}} \\
\texttt{VQA Finetune} & 50.71 & 70.55 \\
\midrule
\rowcolor{gray!10}
\multicolumn{3}{l}{\textbf{Self-Training Methods}} \\
\texttt{STaR\cite{zelikman2022star}} & 55.92 & 72.09  \\
\texttt{R3V\cite{cheng2025vision}} & 52.13 & 71.42 \\
\texttt{\textbf{Ours}} & 60.67 & 72.52 \\
\bottomrule
\end{tabular}
\end{scriptsize}
\label{tab:generalize_llava_methods}
\end{minipage}

\end{table}

%% file: figures/qual_1.tex
\begin{figure}[t]
\centering
  \includegraphics [width=0.99\textwidth]{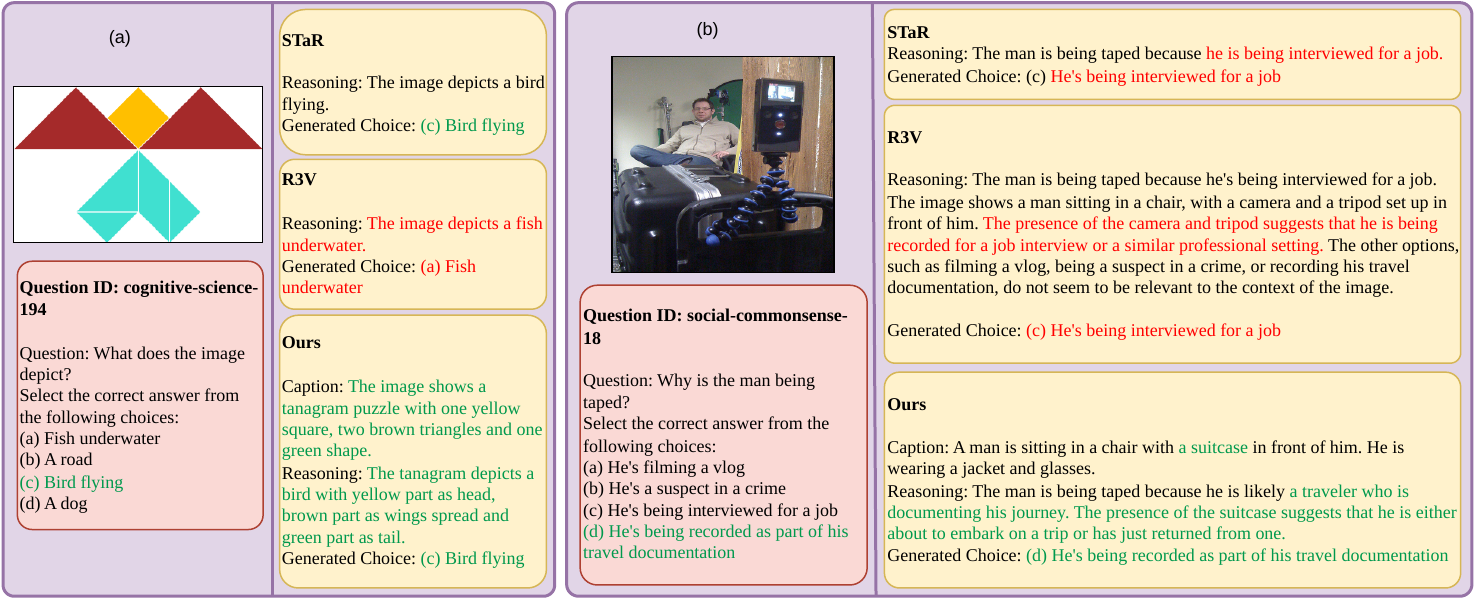}
\caption{\textbf{Qualitative Analysis.} (a) a test example where our method produces the correct answer with high-quality caption and reasoning, while the baselines, STaR \cite{zelikman2022star} and R3V \cite{cheng2025vision}, generate less detailed and visually hallucinated rationales (e.g., fish underwater). (b) An example of a language shortcut (camera -> job interview) in STaR and R3V rationales. Our method correctly identifies key objects such as the suitcase, which grounds the reasoning and produces the correct answer.}
\label{qual4}
\end{figure}

%% file: figures/joint.tex
\begin{figure}[t]
\centering

\begin{minipage}{0.48\linewidth}
\centering
\includegraphics[width=\linewidth]{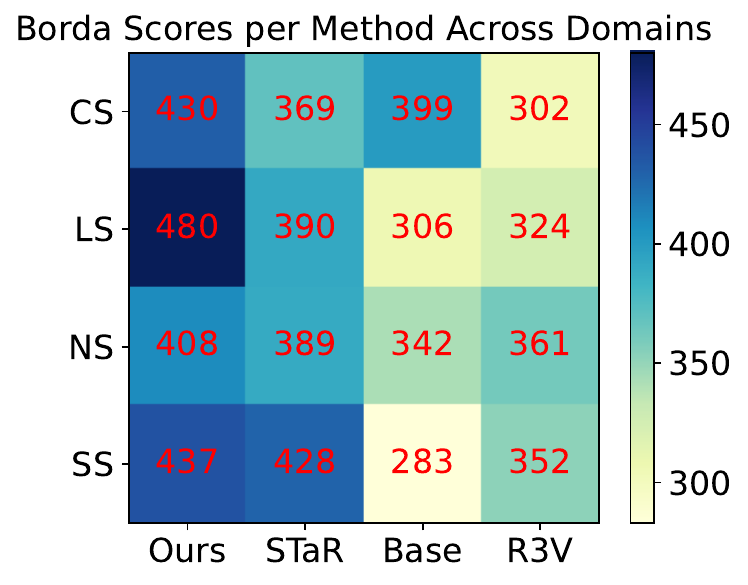}
\caption{\textbf{Subjective Analysis.} Our generated rationales are preferred more often, indicating better quality.}
\label{fig:heatmap}
\end{minipage}
\hfill
\begin{minipage}{0.48\linewidth}
\centering
\includegraphics[width=0.8\linewidth]{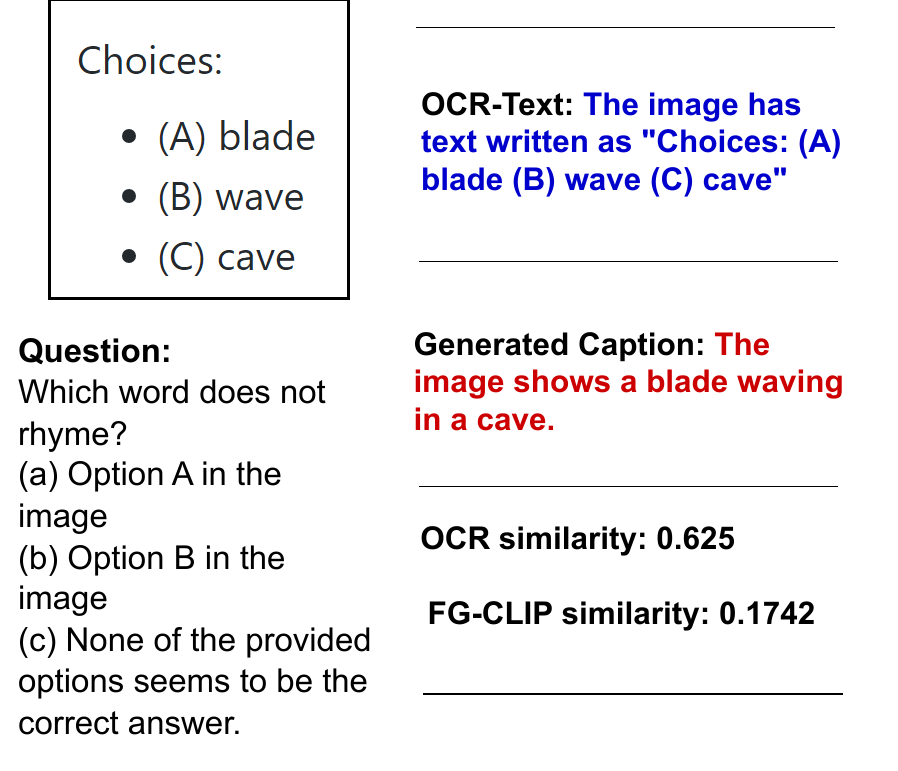}
\caption{Importance of FG-CLIP in preventing failure cases for text-dominated images.}
\label{textfig}
\end{minipage}

\end{figure}

%% file: tables/hallucination.tex
\begin{table}[t]
\centering
\caption{Accuracy (\%) on the OCR split of HallusionBench~\cite{Guan_2024_CVPR}. All models are trained on the Language Science domain of M3CoT~\cite{chen-etal-2024-m3cot}.}
\setlength{\tabcolsep}{7pt}
\renewcommand{\arraystretch}{1.1}
\begin{scriptsize}
\begin{tabular}{lccccc}
\toprule
\textbf{Metric} & \texttt{Base} & \texttt{Direct SFT} & \texttt{STaR} & \texttt{R3V} & \texttt{\textbf{Ours}} \\
\midrule
\textbf{Accuracy (\%)} & 37.76 & 48.25 & 54.54 & 55.24 & \textbf{60.14} \\
\bottomrule
\end{tabular}
\end{scriptsize}
\label{tab:hallusionbench_ocr}
\end{table}

%% file: tables/joint_ablation.tex
\begin{table}[t]
\centering

\begin{minipage}{0.46\linewidth}
\centering
\caption{Ablation study showing the effect of caption filtering and curriculum learning on VQA performance (\%).}
\setlength{\tabcolsep}{2pt}
\renewcommand{\arraystretch}{1}
\begin{scriptsize}
\begin{tabular}{lcc}
\toprule
\textbf{Cases} & \textbf{Lang. Sci.} & \textbf{Common.} \\
\midrule
\texttt{Ours-Stage1 (Ans Filter)} & 49.76 & 62.42 \\
\texttt{Ours-Stage1 (Dual Filter)} & 60.66 & 69.45 \\
\texttt{Ours (W/O CL)} & 57.34 & 65.05 \\
\texttt{Ours} & \textbf{64.93} & \textbf{72.74} \\
\bottomrule
\end{tabular}
\end{scriptsize}
\label{tab:ablation_cot_caption}
\end{minipage}
\hfill
\begin{minipage}{0.46\linewidth}
\centering
\caption{Ablation on PerceptEval components: OCR and FG-CLIP similarity.}
\setlength{\tabcolsep}{1pt}
\renewcommand{\arraystretch}{1}
\begin{scriptsize}
\begin{tabular}{lcc}
\toprule
\textbf{Cases} & \textbf{Lang. Sci.} & \textbf{Common.} \\
\midrule
\texttt{Only OCR} & 61.13 & 60.44 \\
\texttt{Only FG-CLIP} & 56.87 & 69.89 \\
\texttt{Ours} & \textbf{64.93} & \textbf{72.74} \\
\bottomrule
\end{tabular}
\end{scriptsize}
\label{tab:ablation_ocr_fgclip}
\end{minipage}

\end{table}

%% file: figures/graph.tex
\begin{figure}[t]
    \centering
    \includegraphics[width=0.95\textwidth]{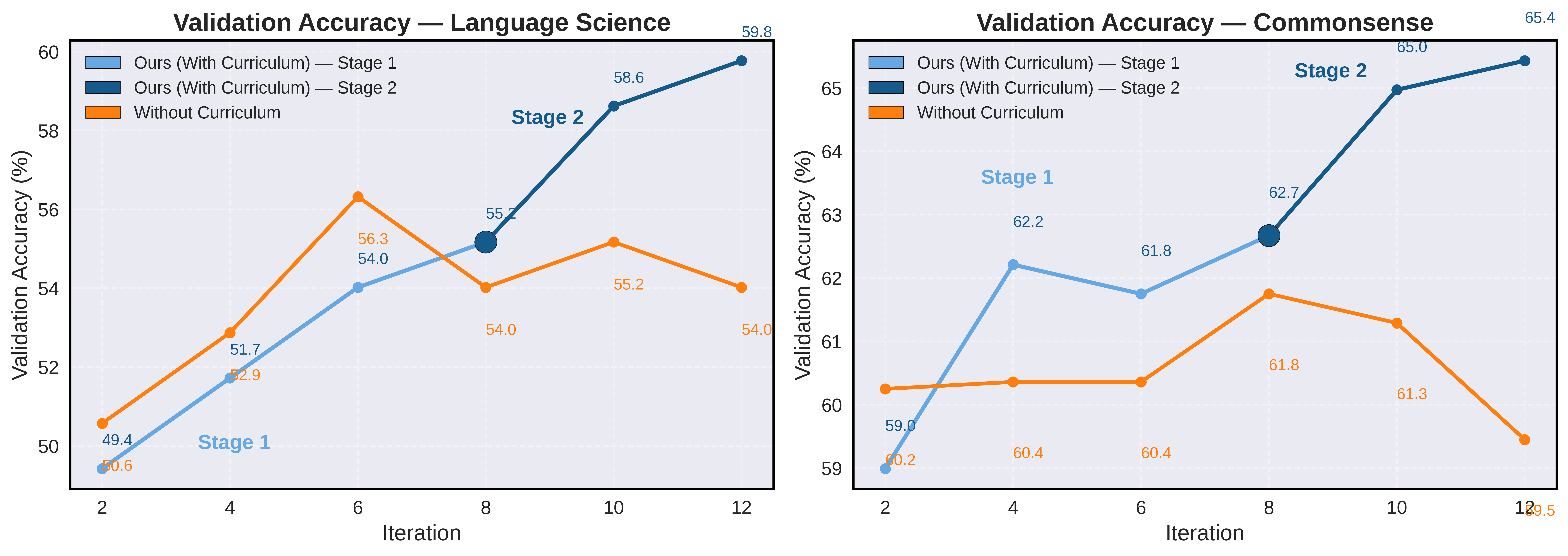}
    \caption{\textbf{Validation accuracy comparison of our method with and without curriculum learning across iterations.} Curves with curriculum learning are shown in blue, split into Stage 1 (light blue) and Stage 2 (dark blue), illustrating the performance progression. The trend without curriculum learning is shown in orange.}
    \label{fig:validation_accuracy}
\end{figure}

%% file: sections/5_concl.tex
\section{Conclusion} \label{sec:conclusions}
Our findings demonstrate that explicitly separating perception from reasoning, coupled with rigorous dual filtering, significantly enhances performance in self-trained VLMs. The proposed PerceptEval metric provides a reliable way to assess caption fidelity without ground truth. Our curriculum strategy further stabilizes training and strengthens reasoning quality across domains. Overall, the framework offers a scalable, low-cost path toward more accurate and visually faithful VLM reasoning.

\section*{Acknowledgement}
This work was supported in part by the Anusandhan National Research Foundation (ANRF) under Prime Minister Early Career Research Grant (PMECRG) Program (Grant No. ANRF/ECRG/2025/005682/ENS).

%% file: sections/X_suppl.tex
\clearpage
\setcounter{section}{0}
\makesupptitle

\title{Improving Reasoning in Vision-Language Models via Perception Verified Self-Training\\[0.5em]
\large Supplementary Material}

\section{PerceptEval Algorithm}
\input{algorithms/percepteval}

\section{Implementation Details}
We provide additional implementation details in this section.

\subsection{Prompt Templates}

\input{figures/prompts}

In this section, we present the prompt templates used in our proposed method. Figure~\ref{prompts} (a) shows the \textit{structured rationale prompt} at the core of our framework. This prompt explicitly disentangles perception from reasoning, facilitating easier optimization while ensuring that the reasoning is grounded in the visual content. It consists of three components:

\begin{itemize}
    \item \textsc{Caption}: Instructs the model to generate a detailed description of the image, including both visual elements and any transcribed text. This caption provides all the relevant information needed to answer the question.

    \item \textsc{Reasoning}: Instructs the model to produce detailed reasoning. Since the reasoning follows the caption, it implicitly relies on the extracted visual details.

    \item \textsc{Conclusion}: Provides the final answer derived from the reasoning process.
\end{itemize}

Figure~\ref{prompts} (b) illustrates the \textit{caption-embedded prompt} used to generate medium-difficulty samples in Stage~2 of our method. For responses where the caption is correct but the reasoning is incorrect, the underlying issue often lies in the model's failure to sufficiently attend to the high-quality caption while reasoning. The caption-embedded prompt addresses this by explicitly conditioning the reasoning on the caption: the prompt begins with the caption itself, followed by instructions to generate the reasoning and conclusion.

When the newly generated conclusion is correct, we pair this refined reasoning with the original accurate caption to construct improved \textsc{[Cap--Reas--Concl]} instances. This augmentation expands the overall training data with moderately difficult samples, increasing the model's robustness. 

\subsection{Dataset Details}
We evaluate our method across four domains, namely commonsense, natural science, language science, and social science, using samples from the M3CoT dataset \cite{chen-etal-2024-m3cot}, which provides multi-domain, multiple-choice visual question-answering. Each domain includes train, validation, and test splits.

The commonsense domain assesses reasoning about physical, social, and temporal aspects depicted in images; natural science focuses on visually grounded questions in physics, chemistry, and biology; social science addresses topics related to geography, economics, and cognitive science; and language science encompasses questions involving figurative language, grammar, and reading comprehension.

Since our proposed caption validation algorithm, \textsc{PerceptEval}, takes into account the content characteristics of an image, we analyse the nature of images across the different domains. The \emph{Language Science} domain predominantly contains text-heavy images, making them largely text-dominated. The \emph{Commonsense} domain, in contrast, features real-world scenes and everyday objects, resulting in primarily visual-dominated images. The \emph{Natural Science} domain typically includes option choices embedded as subimages, often accompanied by textual descriptions related to the subimages or the question itself. Finally, the \emph{Social Science} domain contains structured visual content such as graphs, charts, and maps, which combine textual and visual components.

\subsection{Training and Evaluation Setup}
\input{tables/hyperparam}

Table~\ref{tab:hyperparams} summarizes the training configurations used in our experiments. We finetuned the projector layers along with LoRA layers of the backbone LLM. Due to memory constraints, we employed a per-GPU batch size of 4 with 4 gradient accumulation steps, resulting in an effective batch size of 16. To optimize training efficiency, we utilized the FlashAttention implementation, enabling both \texttt{bf16} and \texttt{tf32} flags for LLaVA \cite{liu2023improvedllava}, and only the \texttt{bf16} flag for Qwen \cite{Qwen2VL}. All experiments were performed on a single NVIDIA A6000 GPU with 48~GB of VRAM.

Some domains in M3CoT \cite{chen-etal-2024-m3cot} exhibit an imbalanced distribution of correct choices. In our experiments, we observed that LVLMs like LLaVA \cite{liu2023improvedllava} tend to have an inherent bias towards option \textbf{A}. When this model bias aligns with the dataset bias, i.e. when most correct answers are also option \textbf{A}, the self-training framework fails to generate meaningful reasoning, as the model can take a shortcut by always predicting \textbf{A} regardless of the rationale. This undermines the iterative process and can ultimately cause the method to collapse. To address this issue, we balance the final choice distribution in the generated rationale dataset according to the correct answer distribution in the full domain training set.

For each domain, we monitored the accuracy on the corresponding validation set to detect saturation for the iterative self-training algorithms. During inference, we set the temperature 0 and limited the maximum number of model tokens to 512. For \textsc{PerceptEval}, we have utilized \textit{fg-clip-large} variant of \texttt{FG-CLIP}, with max length set to 248 enabling long caption validation and \textit{all-MiniLM-L6-v2} variant of \textsc{Sentence-Transformers} for embedding generation.

In the original STaR~\cite{zelikman2022star} framework, few-shot prompting is used to reduce noise in the generated bootstrapped rationale set. Likewise, R3V~\cite{cheng2025vision} mitigates such noise by building the implementation and evaluation pipeline on a GPT-distilled baseline, where GPT-4o supplies high-quality CoT rationales for a small subset of each dataset. In contrast, we do not use few-shot prompting for STaR in our comparisons, nor do we adopt any GPT-based distillation as in R3V. This choice stems from the limited few-shot prompting capabilities of LVLMs and our core principle that \textit{quality outweighs quantity}: careful selection of high-quality rationales is more effective than relying on expensive GPT-generated warm-up annotations for each domain.

\subsection{Subjective Analysis Setup}
\input{figures_supp/subjective}
To evaluate reasoning quality subjectively, we follow the STaR \cite{zelikman2022star} setup. For each M3CoT \cite{chen-etal-2024-m3cot} domain, we sample 50 instances where all methods produce the correct answer. Each instance includes four anonymized rationales, which are ranked by three experts based on reasoning quality (1=best, 4=worst). Rankings are aggregated using Borda score and visualized as a heat map. We have compared rationales generated from our method with those generated through STaR \cite{zelikman2022star}, R3V \cite{cheng2025vision} and zero-shot CoT \cite{letsthinkstepbystep} from base model \textit{LLaVA-v1.5-7B} \cite{liu2023improvedllava}. For our method, we have shown only reasoning part from our generated rationale, appended with "Therefore, the correct answer is {conclusion}.". This is done to ensure the anonymity. Figure~\ref{fig:subjective_gui} illustrates the interface developed to collect these annotations.

\section{Additional Quantitative Results}



\input{tables/output_length}
\input{tables/supp_additional}
\input{tables/supp_llava_next}

\noindent \textbf{Detailed Token Analysis.}
Table~\ref{tab:token_length} reports the detailed token efficiency analysis presented in the main paper. Our method produces responses of moderate length (69.2--93.9 tokens), closely matching the base model (63.6--105.0 tokens), thereby balancing explicit reasoning with token efficiency.

\noindent \textbf{Generalization to LLaVA-NeXt.}
To evaluate the generality of our framework across architectures, we further conduct experiments using the \textit{LLaVA-NeXt} \cite{liu2024llavanext} backbone. On the Language Science (LS) domain, the base model achieves 51.18\% accuracy, while R3V improves performance to 63.98\%. Our method further increases accuracy to 72.04\%, demonstrating that the proposed perception-verified self-training framework generalizes to modern multimodal LLM architectures.

\noindent \textbf{Effect of Training on the Curated Subset.}
To examine whether the observed improvements arise solely from using cleaner training data, we perform direct supervised fine-tuning, i.e. training on the final answer,  on only the samples retained after the final iteration of our framework, \texttt{SFT (curated)} in Table~\ref{tab:additional_baselines}. Despite using the same curated subset, this baseline achieves substantially lower performance than our full method, indicating that the gains originate from the proposed perception-verified self-training procedure rather than dataset filtering alone.

\noindent \textbf{Comparison with DPO}
To evaluate whether preference optimization provides similar benefits under our setting, we replace supervised fine-tuning (SFT) with Direct Preference Optimization (DPO) \cite{dpo} while keeping all other experimental settings unchanged. As shown in Table~\ref{tab:additional_baselines}, DPO \cite{dpo} achieves only 50.71\% accuracy, substantially lower than our proposed method. This suggests that simply replacing SFT with preference optimization is insufficient to realize the benefits of perception-verified self-training under our experimental setting. These findings are in agreement with the observations reported by R3V~\cite{cheng2025vision}.

\noindent \textbf{Effect of Incorporating Hard Samples.}
The proportion of hard samples in the LS domain naturally decreases throughout self-training, from 45.82\% during the first iteration to only 9.70\% in the final iteration. To investigate whether further improvements can be obtained by retaining these challenging samples, we first generate captions using a stronger teacher, \textit{LLaVA-v1.5-13B }\cite{liu2023improvedllava} model, and then apply caption-guided reasoning enhancement. Directly incorporating these samples without curriculum learning achieves 59.24\% accuracy (Table~\ref{tab:additional_baselines}). Introducing them as an additional third stage within our curriculum learning framework improves performance to 65.40\%, only marginally higher than the original method (64.93\%). This suggests that excluding hard samples does not substantially limit performance, while further emphasizing the importance of curriculum learning for stable self-training.

\input{figures_supp/confusion_matrix}

\section{Additional Subjective Analysis}
To further evaluate the reliability of PerceptEval, we conduct a subjective human evaluation to measure its agreement with human judgments. Specifically, we randomly sample 100 captions filtered by PerceptEval from the Language Science (LS) domain, consisting of 50 captions predicted as correct and 50 predicted as incorrect with \textit{LLaVA-v1.5-7B} model. Three human annotators independently assess whether each caption is visually grounded and accurately describes the corresponding image. The final ground-truth label is obtained using majority voting. 

Figure~\ref{fig:confusion_matrix} presents the resulting confusion matrix. The results show that PerceptEval achieves strong agreement with human annotations. Owing to its domain-adaptive confidence threshold, the framework is intentionally conservative, retaining only captions with high confidence while filtering out ambiguous samples. This design substantially reduces false positives, ensuring that captions accepted for self-training are of high quality. Although the stricter threshold leads to a small increase in false negatives, it effectively minimizes noisy supervision, which is more critical for the subsequent self-training process. Overall, these findings demonstrate that PerceptEval provides reliable caption-quality assessment and closely aligns with human judgments.

\section{Additional Qualitative Results}

We present additional qualitative results on the four domains of M3CoT \cite{chen-etal-2024-m3cot}. We compare the responses of self-training methods STaR \cite{zelikman2022star}, R3V \cite{cheng2025vision} and Ours on LLaVA-v1.5-7B \cite{liu2023improvedllava}. Figure \ref{ls1}, Figure \ref{ls2} and Figure \ref{ls3} are from language science domain; Figure \ref{cs1}, Figure \ref{cs2}, Figure \ref{cs3} and Figure \ref{cs4} are from commonsense domain; Figure \ref{ns1}, Figure \ref{ns2} and Figure \ref{ns3} are from natural science domain; Figure \ref{ss1}, Figure \ref{ss2} and Figure \ref{ss3} are from social science domain. As the figures show, both STaR and R3V exhibit visual hallucinations and language biases. This arises because their training pipelines keep perception and reasoning entangled, while supervision is provided only through the final answer. As a result, the models often produce poor-quality rationales even when the final prediction is correct. Moreover, STaR’s hint-based augmentation strategy introduces additional flaws: it encourages shortcut learning, where the model’s final conclusion is no longer based on the rationale it produces. In contrast, our framework explicitly disentangles perception from reasoning and grounds the reasoning process on the generated caption. Furthermore, we provide feedback on both the caption and the final verdict, jointly optimizing these components during training. This dual-level supervision leads to higher–quality rationales and substantially enhances the model's reasoning capabilities.

We visualize the rationales generated on the test sets of Natural Science (Figure~\ref{qual5} (a)) and Commonsense (Figure~\ref{qual5} (b)) tasks, qualitatively comparing our method with existing self-training approaches, STaR and R3V. In Figure~\ref{qual5} (a), we observe that due to the entanglement of reasoning and perception, current methods often produce poor rationales, failing to attend to crucial visual details and resulting in incorrect answers. For instance, STaR misidentifies the orientation of the magnets' poles, leading to confusion and an incorrect conclusion. R3V, while better at outlining a problem-solving approach compared to STaR, still overlooks key visual details, resulting in flawed reasoning and an incorrect answer. In contrast, our method generates rationales that are both visually detailed and logically coherent, leading to correct conclusions. Figure~\ref{qual5} (b) illustrates the quality of correct rationales for STaR, R3V, and our method. While all approaches produce correct answers, our rationales are more detailed, structured, and visually aligned compared to STaR and R3V.

As illustrated in Figure \ref{hallucinations}, the self-training baselines STaR \cite{zelikman2022star} and \cite{cheng2025vision} produce incorrect answers due to hallucinations and linguistic biases, whereas our method arrives at the correct answer by grounding reasoning in accurate visual perception.

\input{figures_supp/ls1}
\input{figures_supp/ls2}
\input{figures_supp/ls3}
\input{figures_supp/cs1}
\input{figures_supp/cs2}
\input{figures_supp/cs3}
\input{figures_supp/cs4}
\input{figures_supp/ns1}
\input{figures_supp/ns2}
\input{figures_supp/ns3}
\input{figures_supp/ss1}
\input{figures_supp/ss2}
\input{figures_supp/ss3}
\input{figures_supp/qual}
\input{figures_supp/mainfig}

\section{Future Scope}
Despite the strengths and promising results of our framework, several directions remain for future work. First, we plan to extend our self-training framework by incorporating DPO \cite{dpo} and comparing it with other RL-based self-training methods. Since these approaches require substantially higher computational resources, such comparisons are beyond the scope of this efficiency-focused study.

Second, our framework currently enforces the \textsc{[Cap-Reas-Concl]} format for every response, even though many questions can be answered directly without detailed reasoning. A promising direction is to develop a \textit{reason-only-if-necessary} mechanism that selectively generates captions and rationales only when beneficial, improving test-time efficiency.

Third, our framework categorizes generated rationales into easy, medium, and hard cases. While hard cases include both partially correct and entirely incorrect predictions, incorporating samples where both the caption and rationale are incorrect introduces considerable noise. Future work will investigate strategies to leverage the more reliable hard-case samples, where the reasoning remains correct despite an incorrect caption, to further improve robustness on unfamiliar inputs.

%% file: algorithms/percepteval.tex
\begin{algorithm}
\caption{\textsc{PerceptEval: Caption Validation via OCR and Visual Agreement}}
\label{alg:perceptev_adaptive_min0}
\begin{algorithmic}[1]
\Require 
    Image $I_i$; Generated caption $\hat{d}_i$; CLIP-based evaluator $\Phi$; 
    OCR module $\Omega$; Sentence-Transformer encoder $f_{\text{st}}$; 
    Max similarity thresholds $\tau_{\max}^{\text{ocr}}, \tau_{\max}^{\text{vis}}$.

\Statex
\State \textbf{(A) OCR–Based Text Agreement}
\State $bboxes \gets \Omega(I_i)$
\State $T_i \gets$ Extract text from $bboxes$
\State $d_i^{\text{ocr}} \gets$ ``The image has text written as: \{$T_i$\}''
\State $s_i^{\text{ocr}} \gets \operatorname{cos-sim}\big(f_{\text{st}}(\hat{d}_i), f_{\text{st}}(d_i^{\text{ocr}})\big)$

\Statex
\State \textbf{(B) Visual Agreement via FG-CLIP}
\State $s_i^{\text{vis}} \gets \Phi(I_i, \hat{d}_i)$

\Statex
\State \textbf{(C) Domain-Aware Joint Evaluation}
\State $\text{r}_t \gets \frac{\sum_{j} \text{area of bbox}_j}{\text{area of image}}$
\State $\tau_{\text{ocr}} \gets \tau_{\max}^{\text{ocr}} \cdot \text{r}_t$ 
\State $\tau_{\text{vis}} \gets \tau_{\max}^{\text{vis}} \cdot (1 - \text{r}_t)$ 

\State $\text{valid}^{\text{cap}}_i \gets (s_i^{\text{ocr}} > \tau_{\text{ocr}}) \land (s_i^{\text{vis}} > \tau_{\text{vis}})$
\State \Return $\text{valid}^{\text{cap}}_i$
\end{algorithmic}
\end{algorithm}

%% file: figures/prompts.tex
\begin{figure}[t]
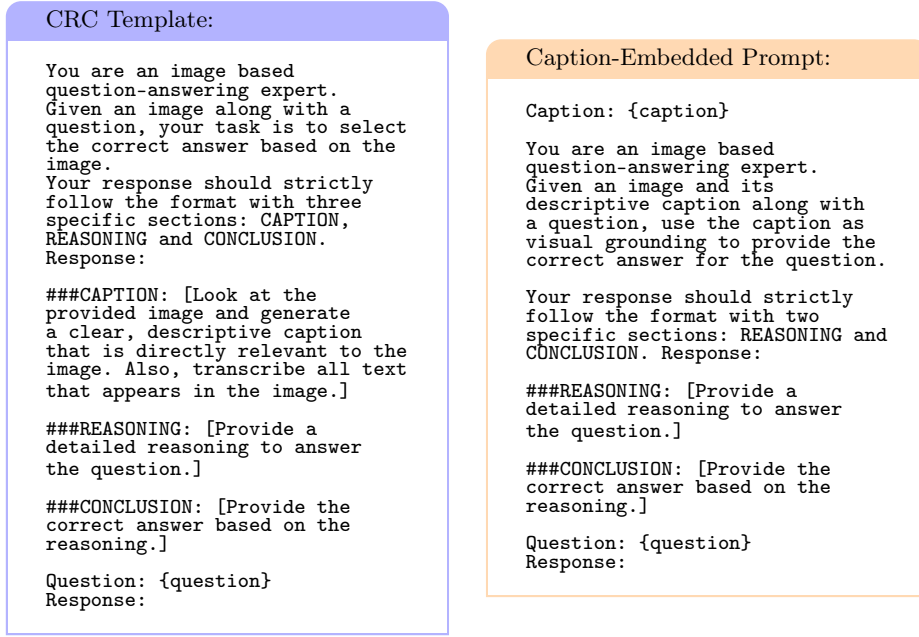

\centering

\begin{minipage}{0.48\textwidth}
\begin{tcolorbox}[
  colback=white,
  colframe=blue!30,
  coltext=black,
  coltitle=black,
  boxrule=0.8pt,
  arc=5pt,
  sharp corners=south,
  fontupper=\fontsize{8}{7}\selectfont\ttfamily,
  title=CRC Template:,
  before skip=5pt,
  after skip=5pt
]
You are an image based question-answering expert.\\
Given an image along with a question, your task is to select the correct answer based on the image.\\
Your response should strictly follow the format with three specific sections: CAPTION, REASONING and CONCLUSION. Response:\\

\#\#\#CAPTION: [Look at the provided image and generate a clear, descriptive caption that is directly relevant to the image. Also, transcribe all text that appears in the image.]\\

\#\#\#REASONING: [Provide a detailed reasoning to answer the question.]\\

\#\#\#CONCLUSION: [Provide the correct answer based on the reasoning.]\\

Question: \{question\}\\
Response:
\end{tcolorbox}
\end{minipage}
\hfill
\begin{minipage}{0.48\textwidth}
\begin{tcolorbox}[
  colback=white,
  colframe=orange!30,
  coltext=black,
  coltitle=black,
  boxrule=0.8pt,
  arc=5pt,
  sharp corners=south,
  fontupper=\fontsize{8}{7}\selectfont\ttfamily,
  title=Caption-Embedded Prompt:,
  before skip=6pt,
  after skip=6pt
]
Caption: \{caption\}\\

You are an image based question-answering expert.\\
Given an image and its descriptive caption along with a question, use the caption as visual grounding to provide the correct answer for the question.\\

Your response should strictly follow the format with two specific sections: REASONING and CONCLUSION. Response:\\

\#\#\#REASONING: [Provide a detailed reasoning to answer the question.]\\

\#\#\#CONCLUSION: [Provide the correct answer based on the reasoning.]\\

Question: \{question\}\\
Response:
\end{tcolorbox}
\end{minipage}

\caption{\textbf{Prompt templates}}
\label{prompts}

\end{figure}

%% file: tables/hyperparam.tex
\begin{table}[t]
\centering
\small
\setlength{\tabcolsep}{1pt}
\begin{tabular}{lcc}
\toprule
\textbf{Hyperparameter} & \textbf{Qwen2-VL} & \textbf{LLaVA-1.5} \\
\midrule
\texttt{Effective Batch Size} & \multicolumn{2}{c}{16} \\
\texttt{LR} & \multicolumn{2}{c}{3e-5} \\
\texttt{Epochs} & \multicolumn{2}{c}{2} \\
\texttt{LR Schedule} & \multicolumn{2}{c}{Constant with warmup} \\
\texttt{LR Warmup Ratio} & \multicolumn{2}{c}{0.1} \\
\texttt{Weight Decay} & \multicolumn{2}{c}{0} \\
\texttt{LoRA Rank} & 64 & 128 \\
\texttt{LoRA alpha} & 16 & 256 \\
\texttt{LoRA Dropout} & \multicolumn{2}{c}{0.05} \\
\texttt{Optimizer} & \multicolumn{2}{c}{AdamW} \\
\bottomrule
\end{tabular}
\caption{Hyperparameter Settings}
\label{tab:hyperparams}
\end{table}

%% file: figures_supp/subjective.tex
\begin{figure}
  \centering
  \includegraphics[width=0.95\textwidth]{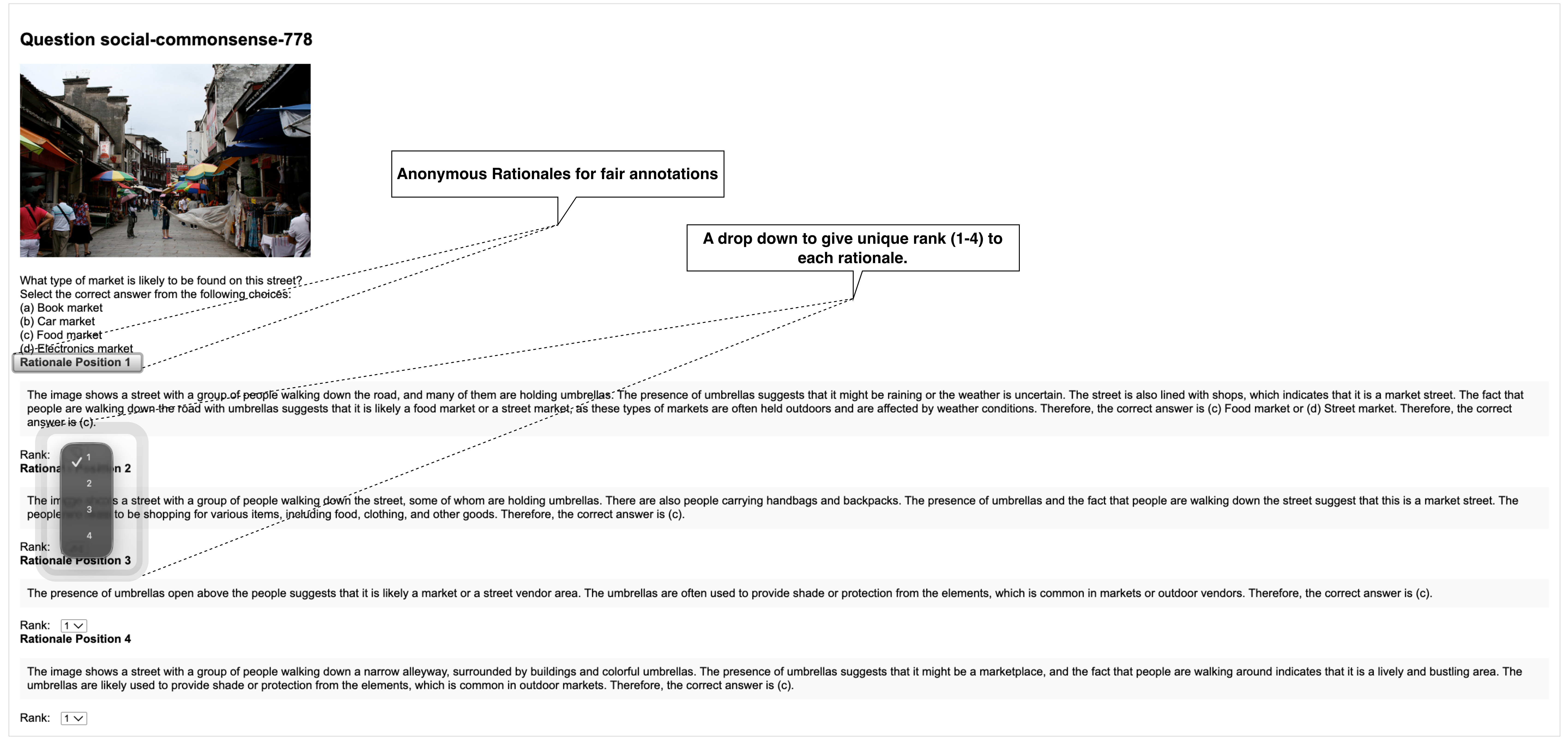}
  \caption{\textbf{Screenshot of our subjective annotation GUI.} Please note that the name of the method has been kept anonymous to ensure a fair comparison.}
  \label{fig:subjective_gui}
\end{figure}

%% file: tables/output_length.tex
\begin{table}[t]
\centering
\caption{Average token length of generated rationales.}
\setlength{\tabcolsep}{4.2pt}
\renewcommand{\arraystretch}{1.1}
\begin{scriptsize}
\begin{tabular}{lcccc}
\toprule
\textbf{Method} & \textbf{Lang. Sci.} & \textbf{Common.} & \textbf{Natural Sci.} & \textbf{Social Sci.} \\
\midrule
\rowcolor{gray!10}
\multicolumn{5}{l}{\textbf{Zero-Shot Method}} \\
\texttt{CoT} & 84.22 & 63.66 & 92.69 & 105.09 \\
\midrule
\rowcolor{gray!10}
\multicolumn{5}{l}{\textbf{Self-Training Methods}} \\
\texttt{STaR\cite{zelikman2022star}} & 31.67 & 52.49 & 56.12 & 49.64 \\
\texttt{R3V\cite{cheng2025vision}} & 65.67 & 80.33 & 100.53 & 126.03 \\
\texttt{\textbf{Ours}} & 79.89 & 69.21 & 93.96 & 87.48 \\
\bottomrule
\end{tabular}
\end{scriptsize}
\label{tab:token_length}
\end{table}

%% file: tables/supp_additional.tex
\begin{table}[t]
\centering
\caption{Additional ablations on the Language Science (LS) domain using \textit{LLaVA-v1.5-7B}.}
\setlength{\tabcolsep}{8pt}
\renewcommand{\arraystretch}{1.1}
\begin{scriptsize}
\begin{tabular}{lc}
\toprule
\textbf{Method} & \textbf{Accuracy (\%)} \\
\midrule
\rowcolor{gray!10}
\multicolumn{2}{l}{\textbf{Baseline Methods}} \\
\texttt{SFT (Curated)} & 45.34 \\
\texttt{DPO} \cite{dpo} & 50.71 \\
\midrule
\rowcolor{gray!10}
\multicolumn{2}{l}{\textbf{Our Variants}} \\
\texttt{Ours + Hard Samples (w/o CL)} & 59.24 \\
\texttt{Ours + Hard Samples (w/ CL)} & \textbf{65.40} \\
\texttt{\textbf{Ours}} & 64.93 \\
\bottomrule
\end{tabular}
\end{scriptsize}
\label{tab:additional_baselines}
\end{table}

%% file: tables/supp_llava_next.tex
\begin{table}[t]
\centering
\caption{Generalization to the \textit{LLaVA-NeXt} architecture on the Language Science (LS) domain.}
\setlength{\tabcolsep}{10pt}
\renewcommand{\arraystretch}{1.1}
\begin{scriptsize}
\begin{tabular}{lccc}
\toprule
\textbf{Methods} & \texttt{Base} & \texttt{R3V} \cite{cheng2025vision} & \texttt{\textbf{Ours}} \\
\midrule
\textbf{Accuracy (\%)} & 51.18 & 63.98 & \textbf{72.04} \\
\bottomrule
\end{tabular}
\end{scriptsize}
\label{tab:llava_next}
\end{table}

%% file: figures_supp/confusion_matrix.tex
\begin{figure}
    \centering
    \includegraphics[width=0.40\linewidth]{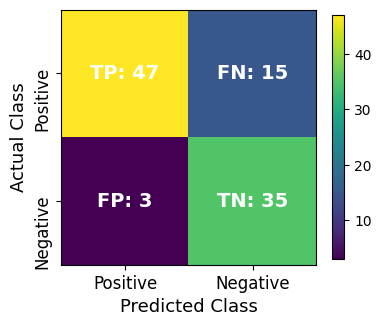}
    \caption{Confusion matrix showing the agreement between PerceptEval and human annotations on the LS domain.}
    \label{fig:confusion_matrix}
\end{figure}

%% file: figures_supp/ls1.tex
\begin{figure*}
\centering
  \includegraphics [width=0.60\textwidth]{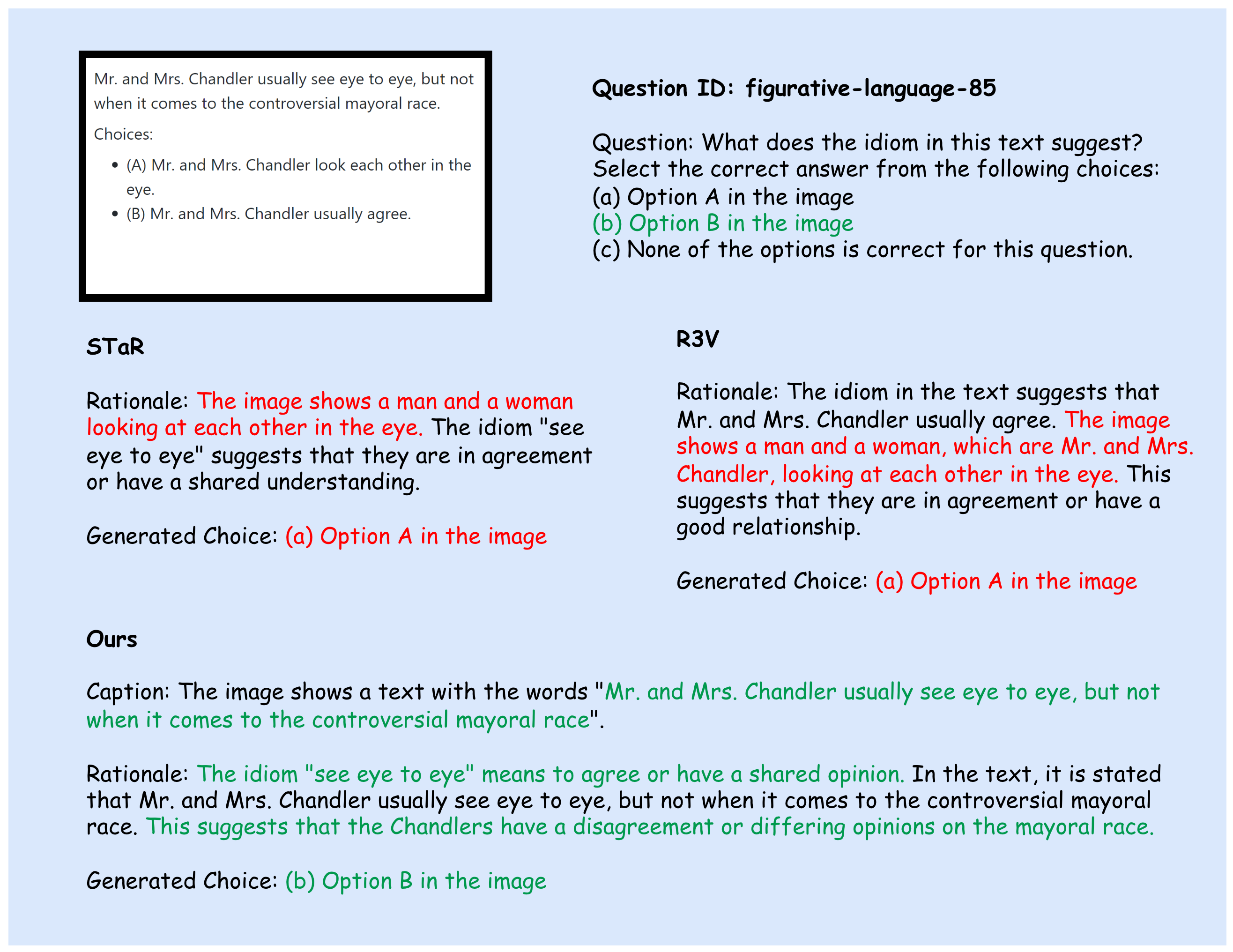}
\caption{\textbf{Qualitative Analysis.} STaR \cite{zelikman2022star} and R3V \cite{cheng2025vision} both hallucinate visual content, \textcolor{red}{misinterpreting the embedded text as an image of a man and woman looking at each other}, which leads to an incorrect answer. In contrast, \textcolor{darkgreen}{our method correctly perceives the textual content and produces the appropriate interpretation}.
}
\label{ls1}
\end{figure*}

%% file: figures_supp/ls2.tex
\begin{figure*}
\centering
  \includegraphics [width=0.60\textwidth]{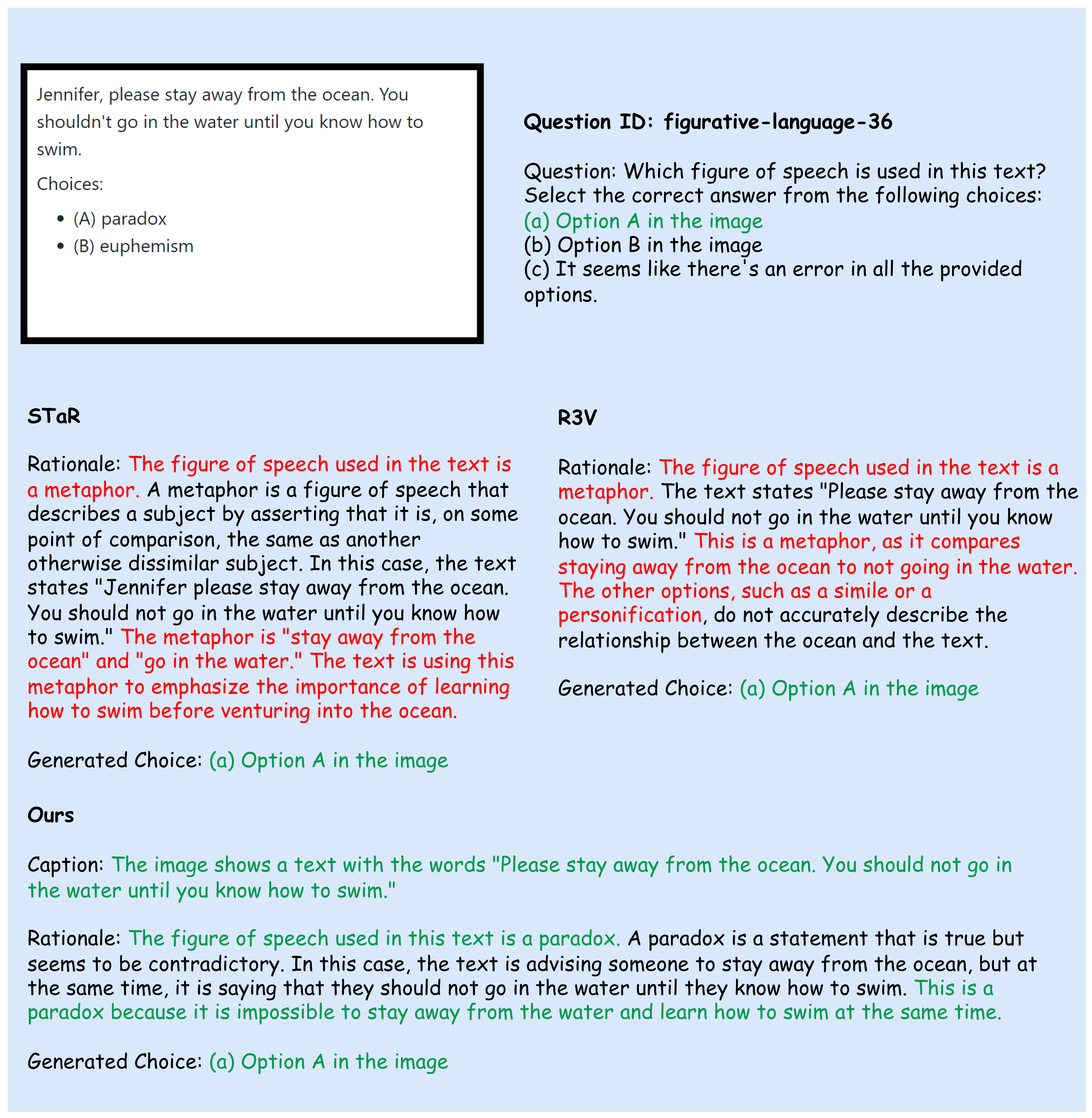}
\caption{\textbf{Qualitative Analysis.} Due to their entangled perception–reasoning pipelines, STaR \cite{zelikman2022star} and R3V \cite{cheng2025vision} amplify the model’s inherent language bias, \textcolor{red}{vaguely linking the figure of speech with metaphor}, leading to incorrect reasoning. Furthermore, the model ended up selecting option A under ambiguity. In contrast, \textcolor{darkgreen}{our method, supported by accurate OCR-based perception, correctly identifies the paradox and generates a detailed, logically sound rationale}.}
\label{ls2}
\end{figure*}

%% file: figures_supp/ls3.tex
\begin{figure*}
\centering
  \includegraphics [width=0.60\textwidth]{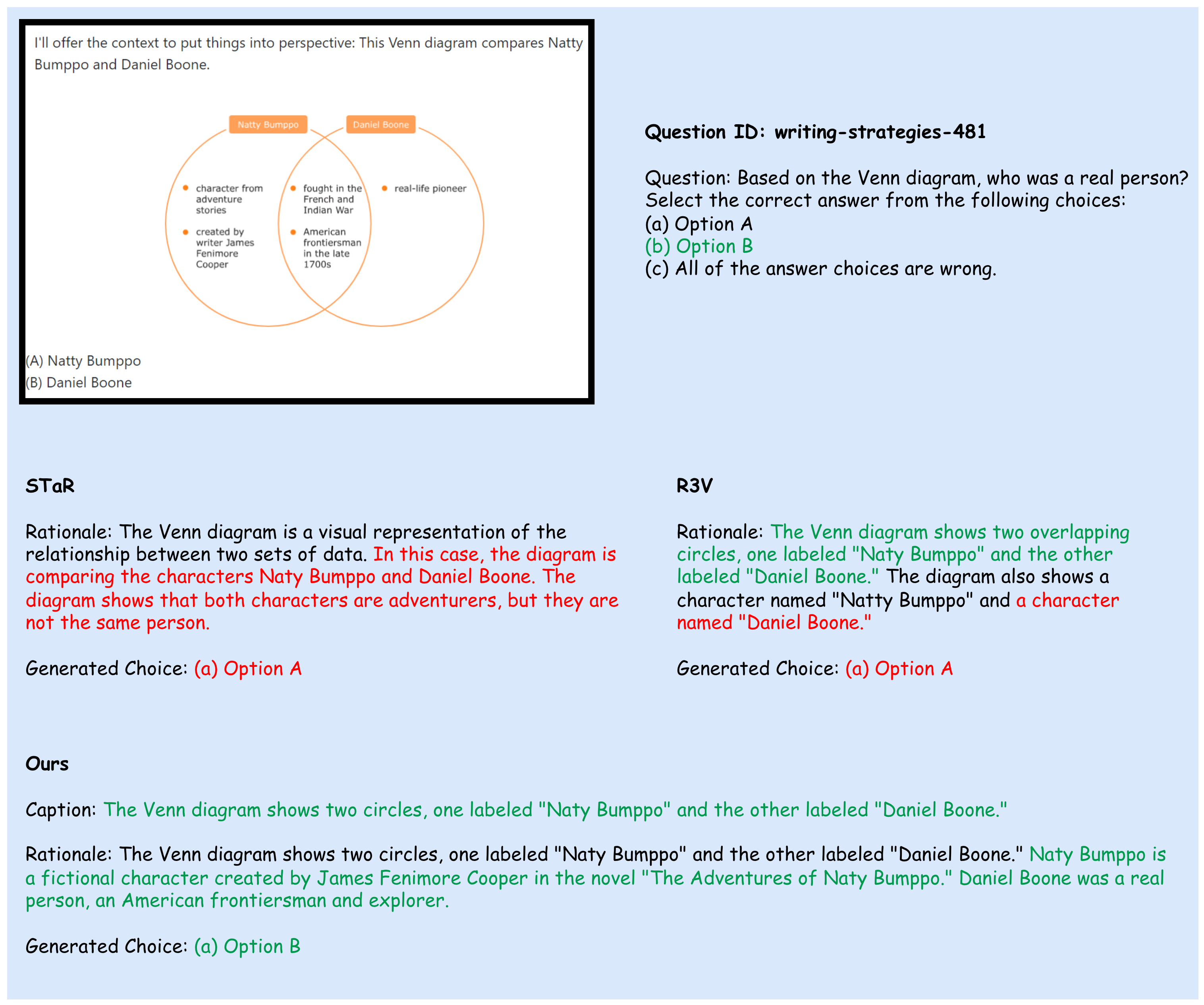}
\caption{\textbf{Qualitative Analysis.} Both STaR \cite{zelikman2022star} and R3V \cite{cheng2025vision} \textcolor{red}{misclassify Daniel Bonne as a character, ultimately resorting to Option A under ambiguity}. \textcolor{darkgreen}{Our method correctly identifies him as a real person due to the correct OCR perception}.}
\label{ls3}
\end{figure*}

%% file: figures_supp/cs1.tex
\begin{figure*}
\centering
  \includegraphics [width=0.6\textwidth]{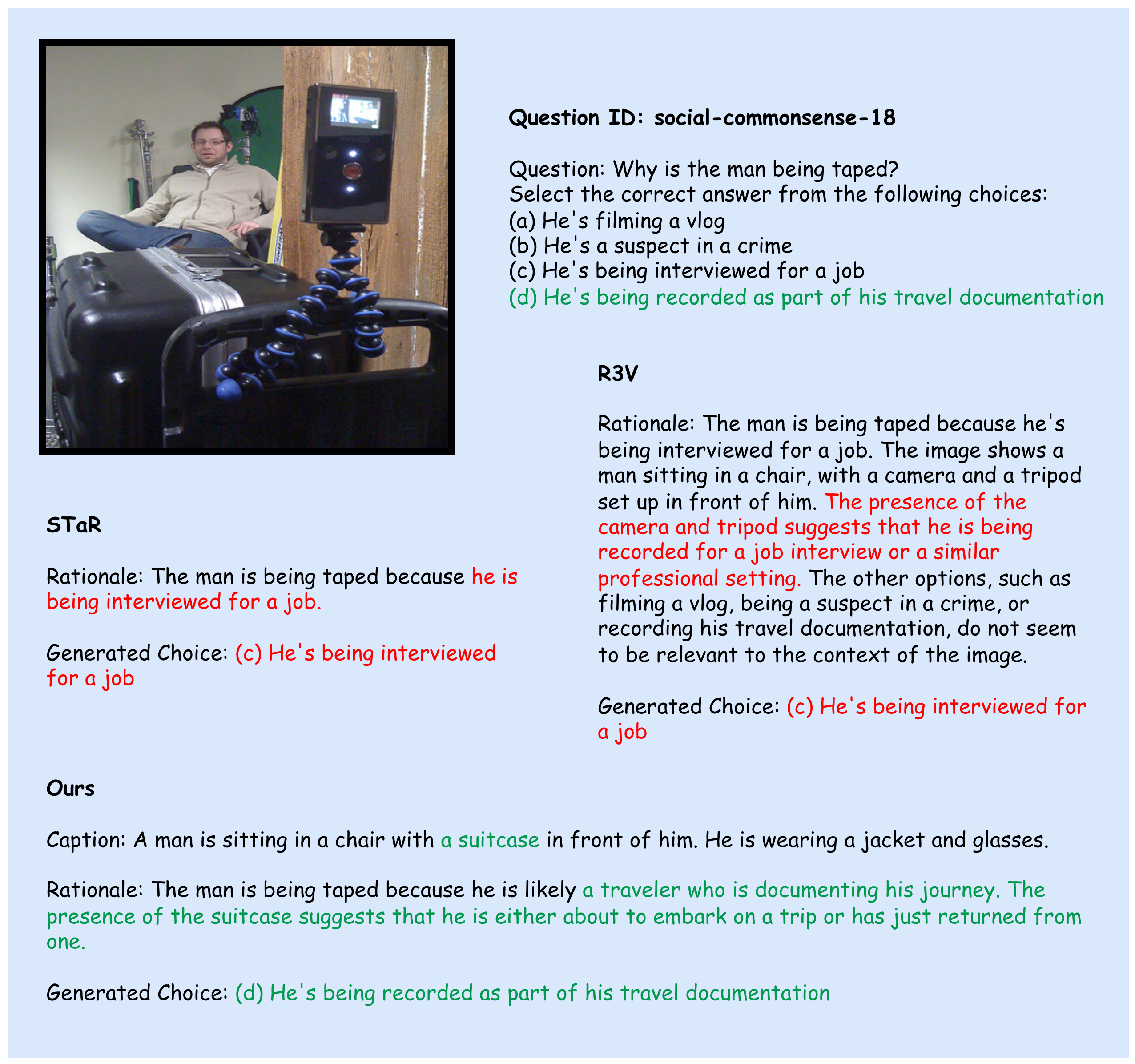}
\caption{\textbf{Qualitative Analysis.} Due to inherent language bias, STaR \cite{zelikman2022star} and R3V \cite{cheng2025vision} \textcolor{red}{misinterpret the scene, associating camera recording with a job interview rather than attending to the visual details}. In contrast, our method correctly identifies key objects such as the \textcolor{darkgreen}{suitcase}, which grounds the reasoning and leads to the correct answer.}
\label{cs1}
\end{figure*}

%% file: figures_supp/cs2.tex
\begin{figure*}
\centering
  \includegraphics [width=0.6\textwidth]{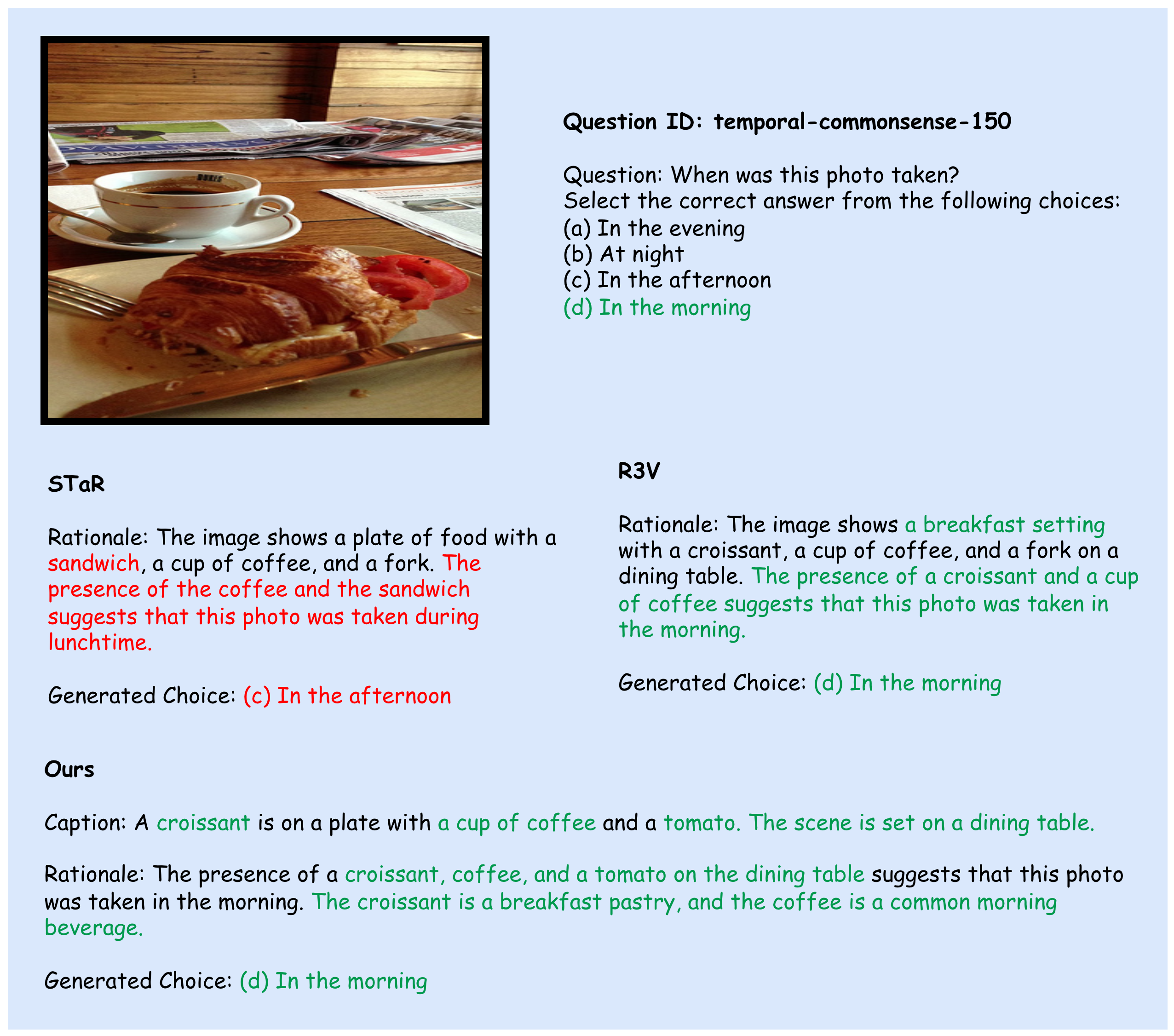}
\caption{\textbf{Qualitative Analysis.} In the figure, STaR \cite{zelikman2022star} \textcolor{red}{incorrectly identifies the croissant as a sandwich}, while R3V \cite{cheng2025vision} and our method identify the key objects correctly, such as \textcolor{darkgreen}{a croissant and a cup of coffee}, leading to correct reasoning. This illustrates the importance of correct perception in sound reasoning.}
\label{cs2}
\end{figure*}

%% file: figures_supp/cs3.tex
\begin{figure*}
\centering
  \includegraphics [width=0.60\textwidth]{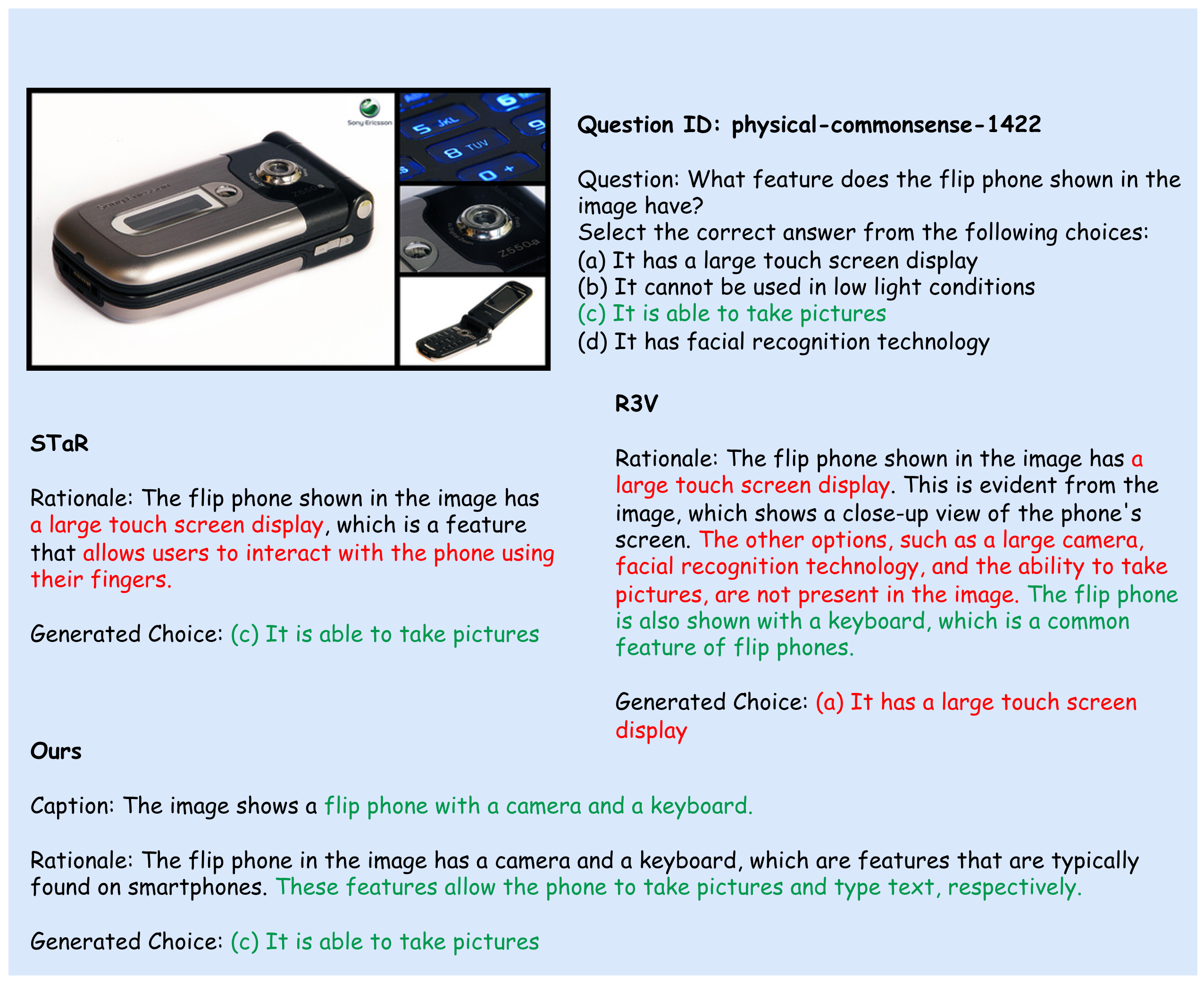}
\caption{\textbf{Qualitative Analysis.} STaR \cite{zelikman2022star} and R3V \cite{cheng2025vision} responses suffer from \textcolor{red}{language bias of associating phone with touch screen display}, hence ignoring the visual details present. Further, \textcolor{red}{due to a flawed hint-based augmentation strategy}, STaR generated the correct answer despite an incorrect rationale, highlighting its tendency to take shortcuts. Our method first correctly identifies the visual details, i.e. \textcolor{darkgreen}{camera and keyboard}, and then conditions the reasoning on the caption.}
\label{cs3}
\end{figure*}

%% file: figures_supp/cs4.tex
\begin{figure*}
\centering
  \includegraphics [width=0.60\textwidth]{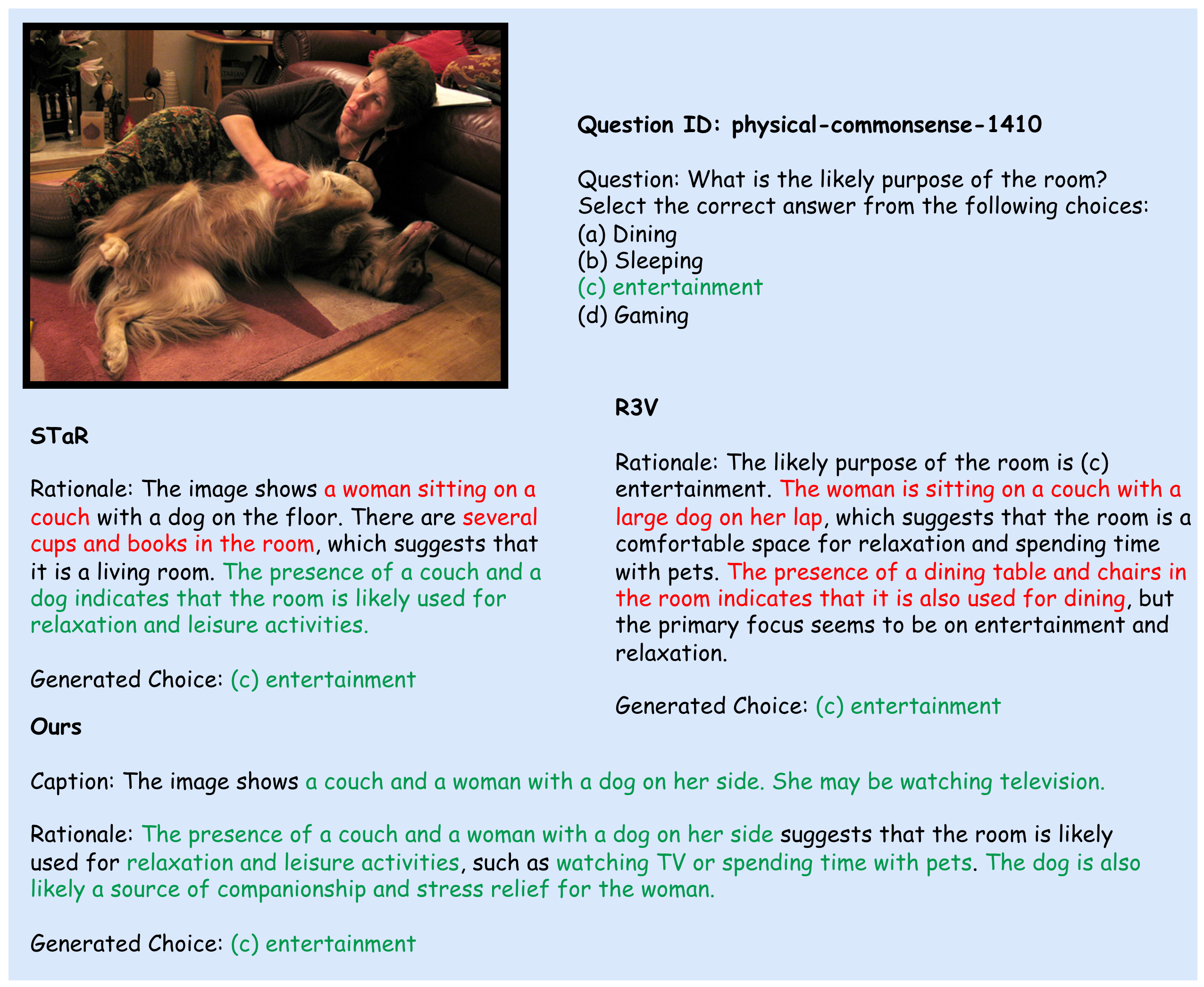}
\caption{\textbf{Qualitative Analysis.} STaR \cite{zelikman2022star} and R3V \cite{cheng2025vision} responses contain visually hallucinating objects such as \textcolor{red}{several cups and books in the room} and \textcolor{red}{dining table and chairs} despite having the correct answer. Our method's rationale correctly attends to visual details, leading to correct interpretation, such as \textcolor{darkgreen}{watching TV or spending time with pets}.}
\label{cs4}
\end{figure*}

%% file: figures_supp/ns1.tex
\begin{figure*}
\centering
  \includegraphics [width=0.60\textwidth]{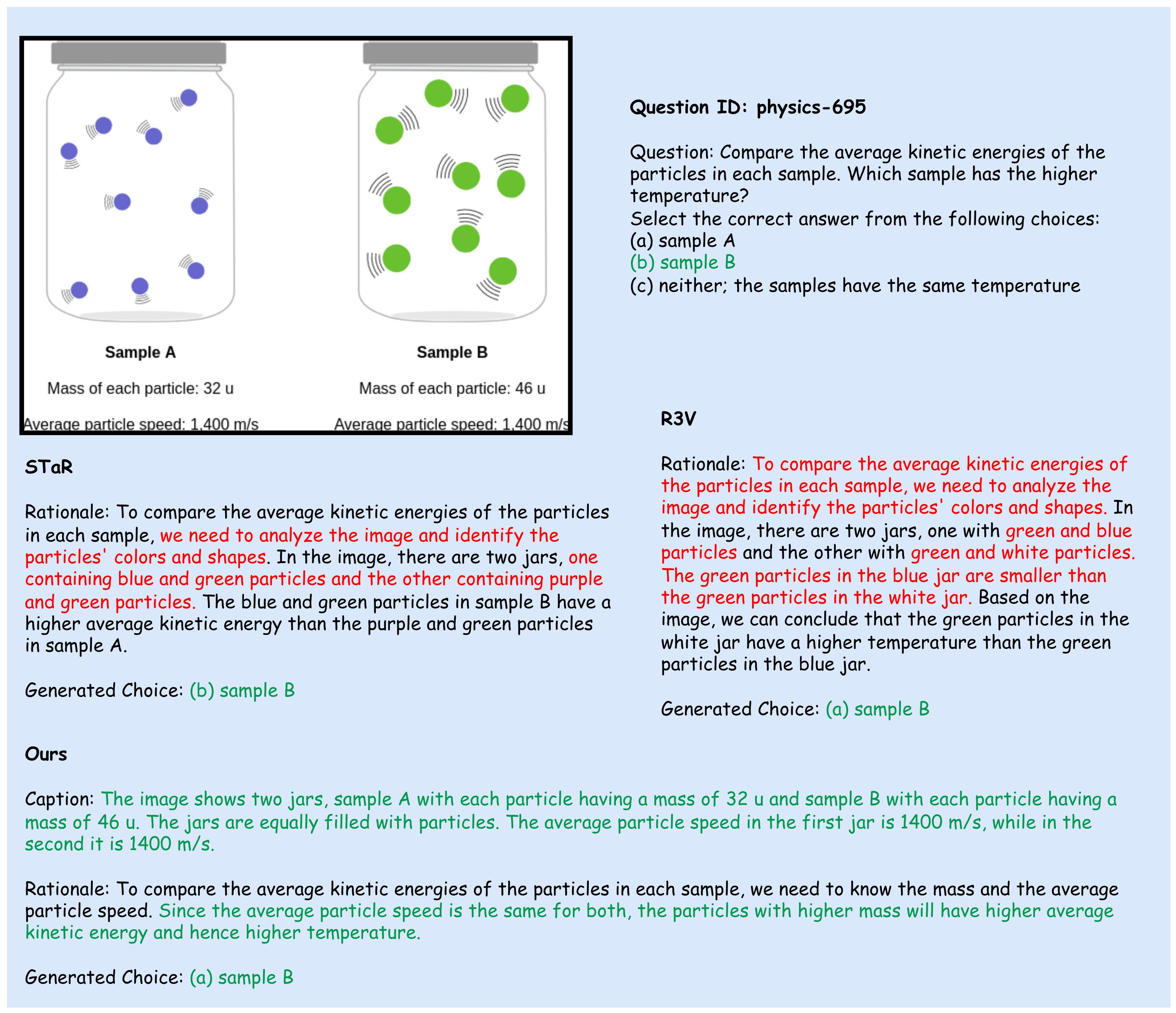}
\caption{\textbf{Qualitative Analysis.} Despite generating correct answers, the rationales of STaR \cite{zelikman2022star} and R3V \cite{cheng2025vision} are highly inconsistent. Firstly, \textcolor{red}{the responses rely on incorrect characteristics, such as colors and shapes of particles, to infer which sample has the higher temperature}, highlighting an incorrect thought process. Secondly, they identify \textcolor{red}{false visual details such as purple particles, green particles in the blue jar}, etc. In contrast, empowered by PerceptEval, our method provides a coherent caption that correctly attends to both textual \textcolor{darkgreen}{(mass and average particle speed)} and visual \textcolor{darkgreen}{(equally filled)} components of the image.}
\label{ns1}
\end{figure*}

%% file: figures_supp/ns2.tex
\begin{figure*}
\centering
  \includegraphics [width=0.60\textwidth]{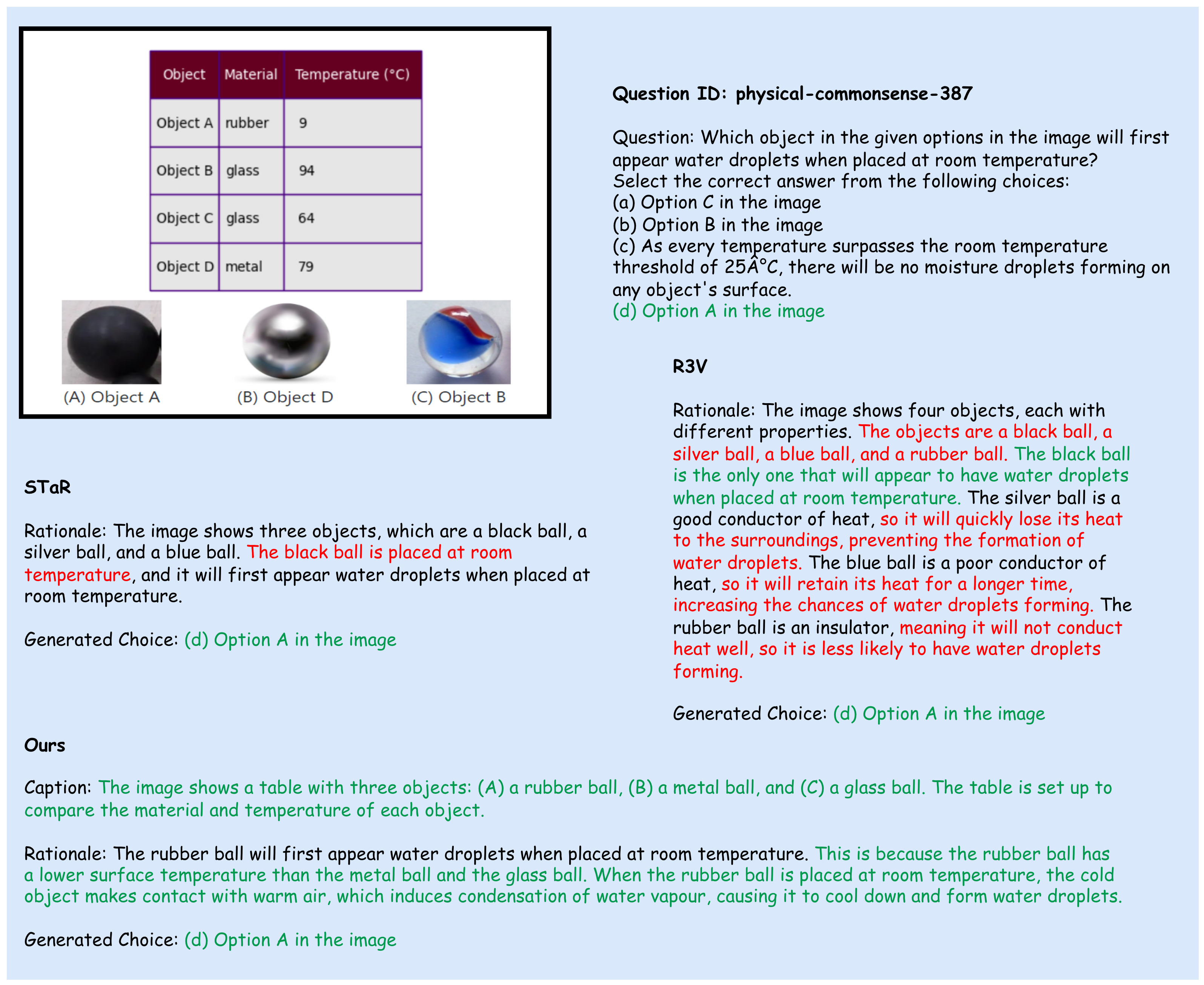}
\caption{\textbf{Qualitative Analysis.} STaR \cite{zelikman2022star} generated insufficient reasoning to correctly answer the question, highlighting its tendency to take shortcuts. Further, R3V \cite{cheng2025vision}, despite generating the correct answer, has flawed reasoning. \textcolor{red}{The response incorrectly relies on the conduction ability of the balls to predict the correct answer, rather than attending to the table present in the figure}. Our method correctly \textcolor{darkgreen}{identifies the role of the table present}. It hence generates sound reasoning, i.e. \textcolor{darkgreen}{based on the transfer of heat between the water vapour and the balls.}}
\label{ns2}
\end{figure*}

%% file: figures_supp/ns3.tex
\begin{figure*}
\centering
  \includegraphics [width=0.70\textwidth]{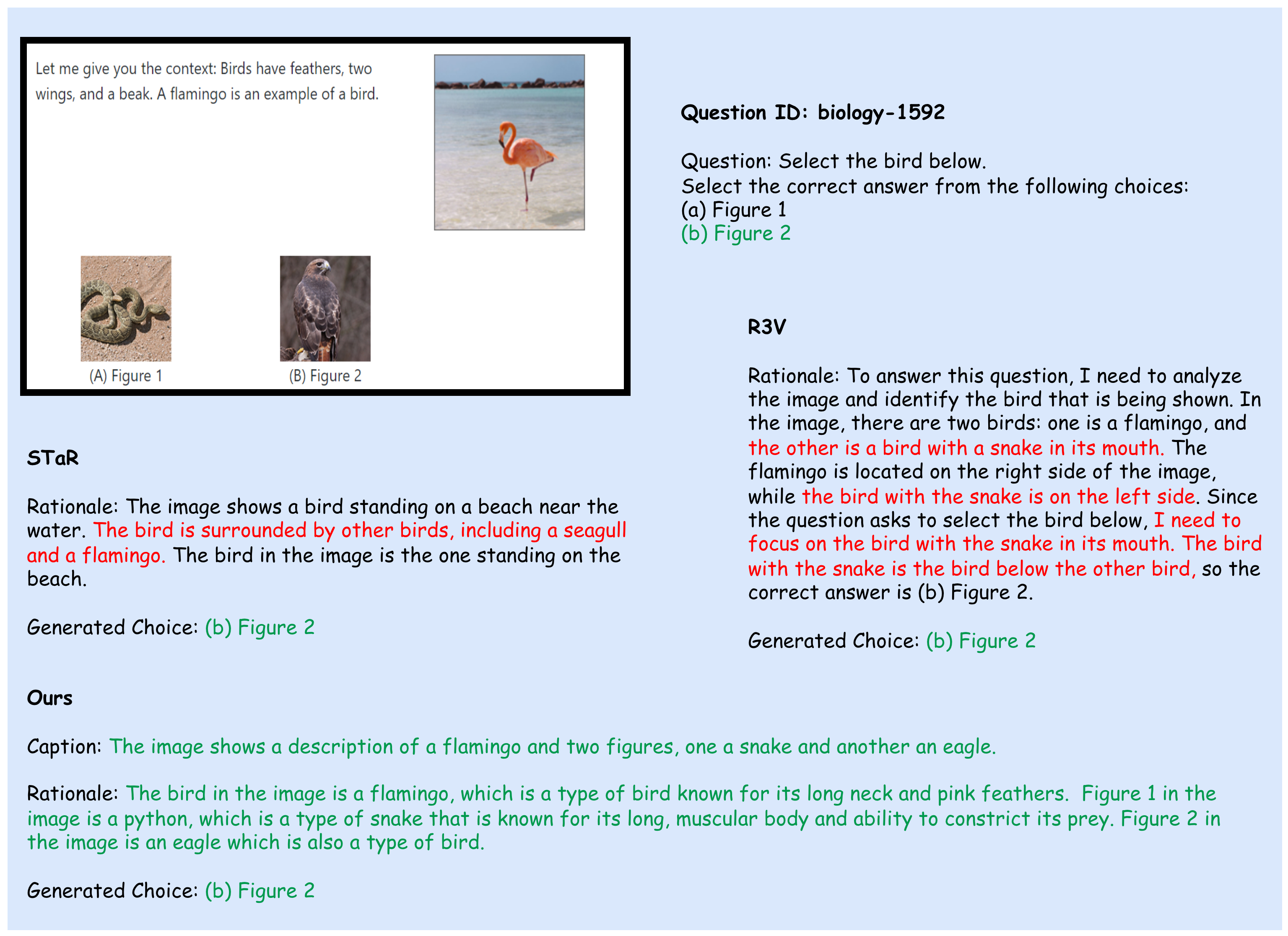}
\caption{\textbf{Qualitative Analysis.} Due to entangled perception and reasoning, STaR \cite{zelikman2022star} and R3V \cite{cheng2025vision} often fall short, producing visual hallucinations, such as \textcolor{red}{a seagull} and \textcolor{red}{a bird with a snake in its mouth}. Due to the correct identification of image components, i.e. \textcolor{darkgreen}{flamingo, description, snake and eagle}, our method's reasoning appears perfect.}
\label{ns3}
\end{figure*}

%% file: figures_supp/ss1.tex
\begin{figure*}
\centering
  \includegraphics [width=0.70\textwidth]{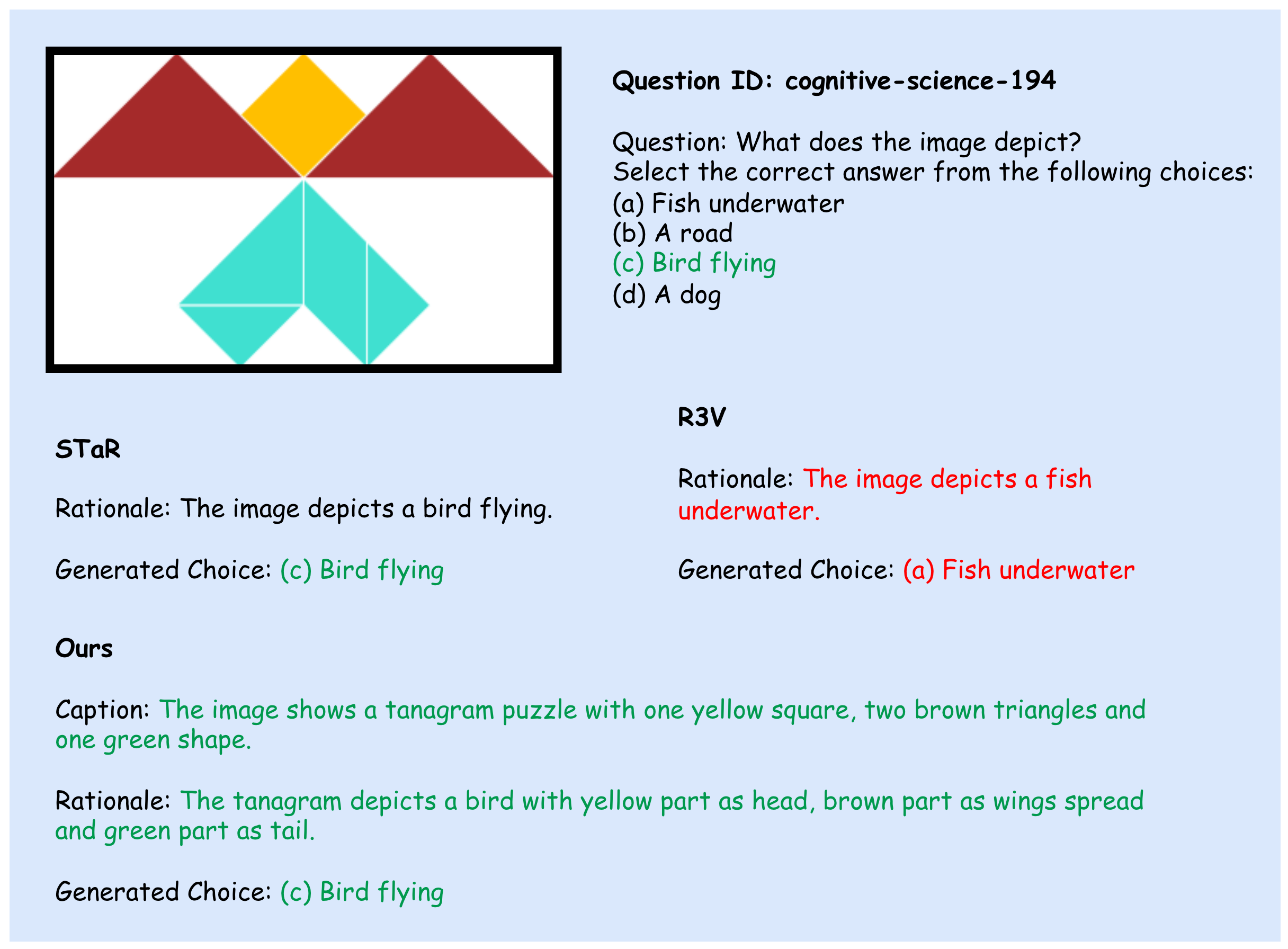}
\caption{\textbf{Qualitative Analysis.} STaR \cite{zelikman2022star} and R3V \cite{cheng2025vision} \textcolor{red}{fail to produce detailed reasoning}, while due to the employed structured rationale prompt, \textcolor{darkgreen}{our method's generated rationale is more detailed and correct}.}
\label{ss1}
\end{figure*}

%% file: figures_supp/ss2.tex
\begin{figure*}
\centering
  \includegraphics [width=0.70\textwidth]{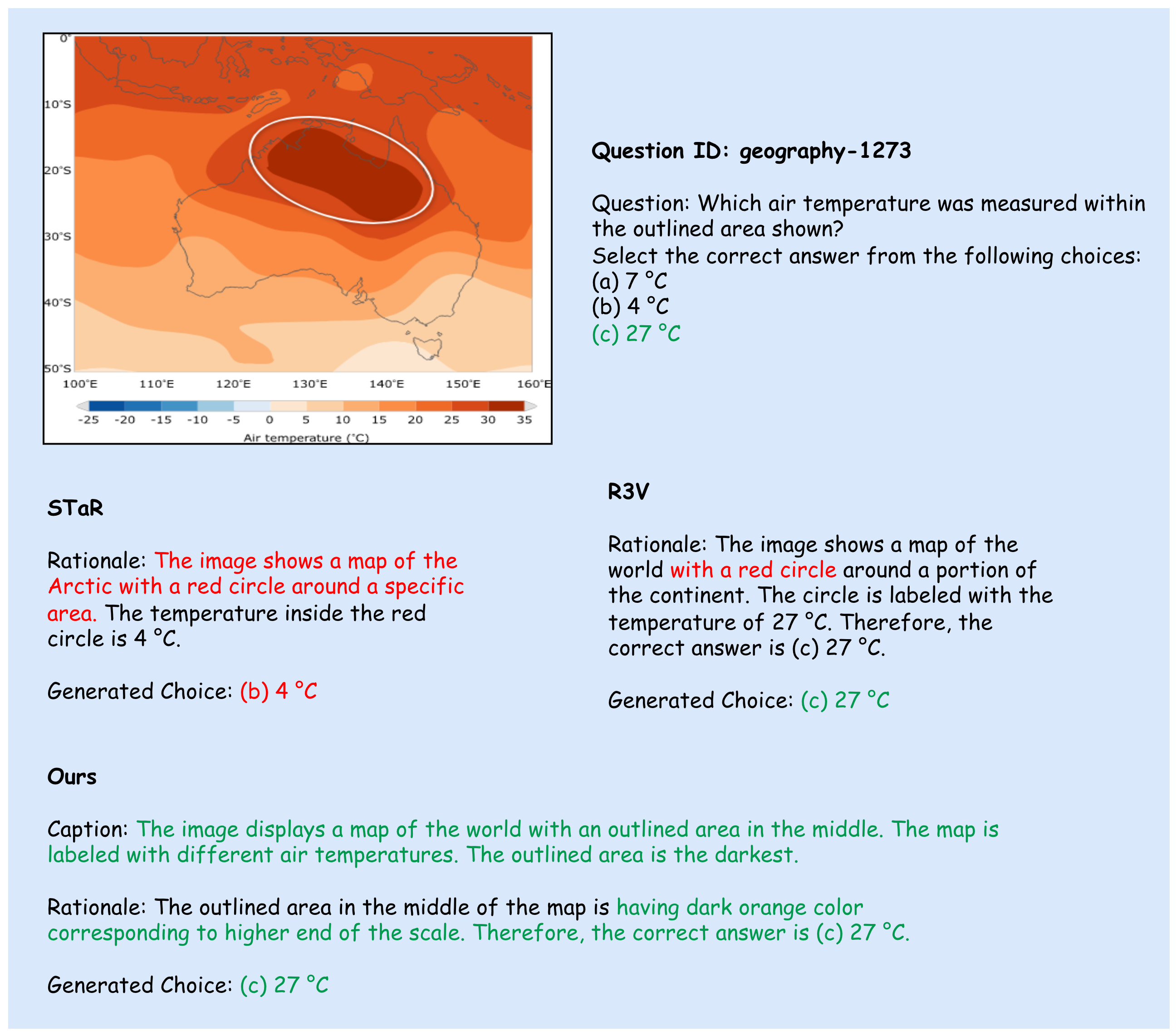}
\caption{\textbf{Qualitative Analysis.} Rationales of both STaR \cite{zelikman2022star} and R3V \cite{cheng2025vision} suffer from visual hallucinations, i.e. \textcolor{red}{Arctic} and \textcolor{red}{red circle}, while our method's rationale is entirely reasonable, i.e. \textcolor{darkgreen}{identifying the temperature from the given temperature scale and color density of the outlined region}.}
\label{ss2}
\end{figure*}

%% file: figures_supp/ss3.tex
\begin{figure*}
\centering
  \includegraphics [width=0.70\textwidth]{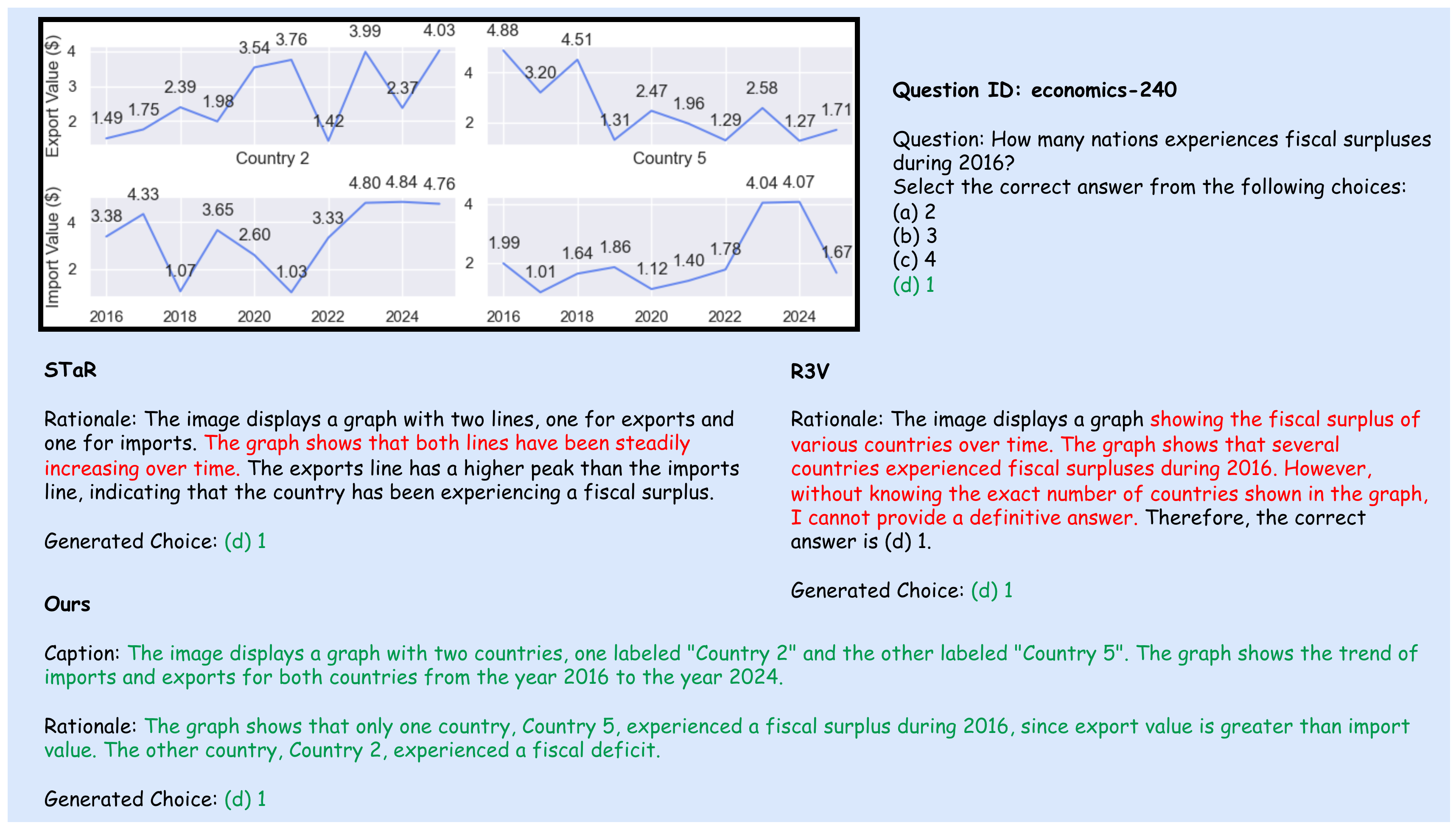}
\caption{\textbf{Qualitative Analysis.} STaR \cite{zelikman2022star} generated \textcolor{red}{inconsistent and insufficient reasoning} to correctly answer the question, while R3V \cite{cheng2025vision} \textcolor{red}{refused to answer the question but guessed the correct answer}. Our method's rationale has \textcolor{darkgreen}{correctly interpreted the trend presented in the graphs to generate the correct answer}.}
\label{ss3}
\end{figure*}

%% file: figures_supp/qual.tex
\begin{figure*}
\centering
  \includegraphics [width=0.99\textwidth]{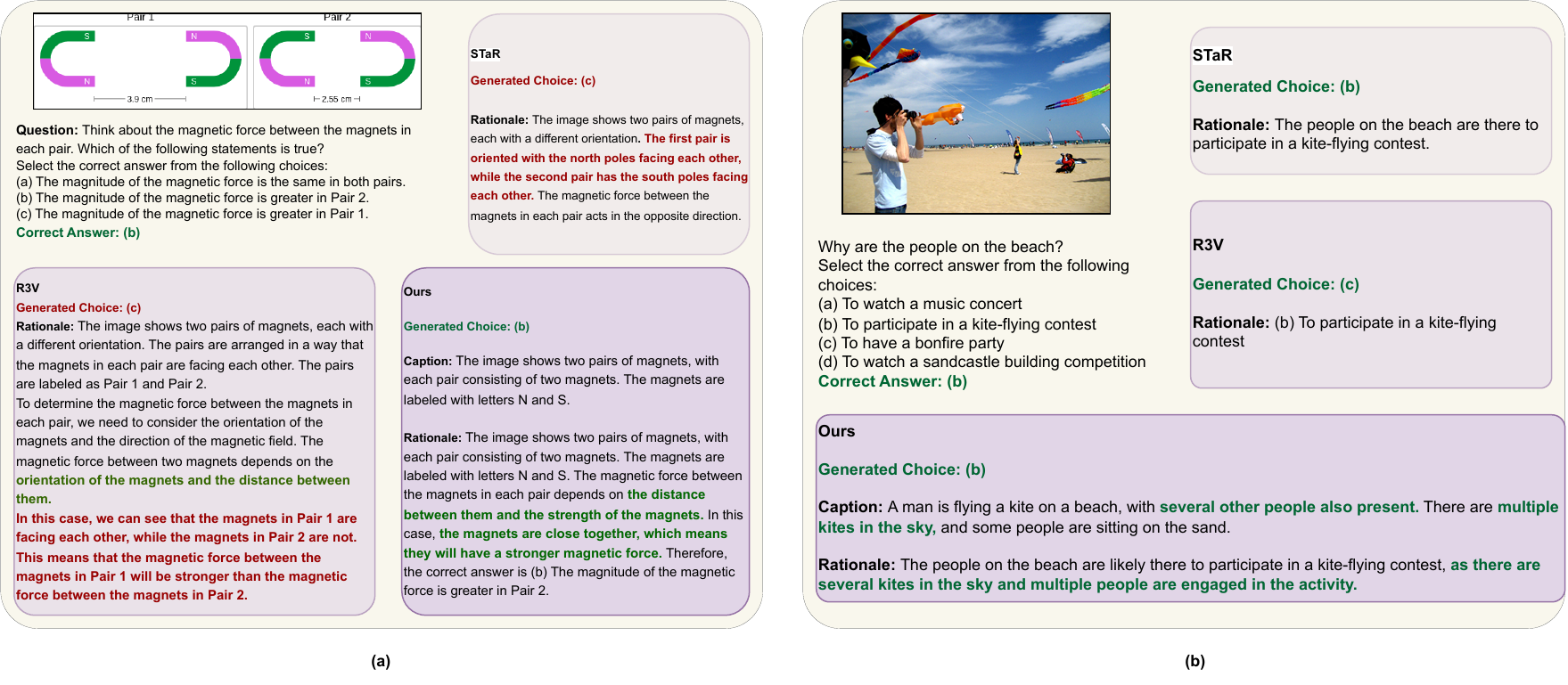}
\caption{\textbf{Qualitative Analysis.} Subfigure (a) presents a test example where our method produces the correct answer along with high-quality captions and reasoning, whereas the baseline methods, STaR \cite{zelikman2022star} and R3V \cite{cheng2025vision}, fail to do so. Subfigure (b) illustrates a case where all methods correctly predict the answer, but our approach generates noticeably more accurate captions and more coherent reasoning. }
\label{qual5}
\end{figure*}

%% file: figures_supp/mainfig.tex
\begin{figure}
\centering
  \includegraphics [width=0.90\textwidth]{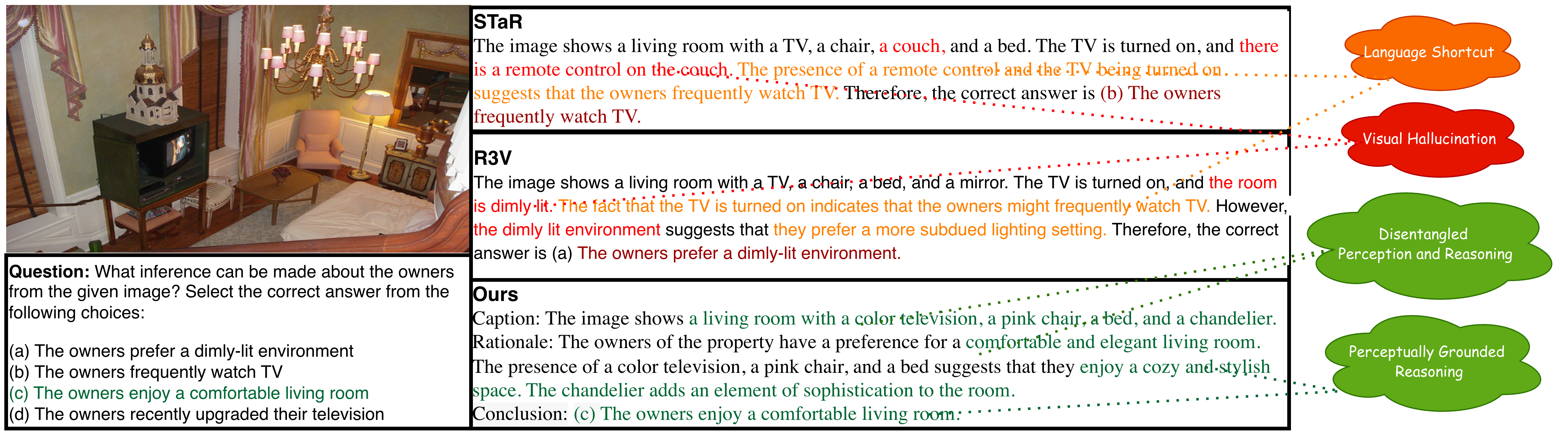}
\caption{\textbf{Comparison with STaR \cite{zelikman2022star} and R3V \cite{cheng2025vision}}. Existing self-training methods for VLMs often struggle with visual hallucinations (e.g., misidentifying objects) and language shortcuts (e.g., relying on biases such as “TV → remote”). Our framework mitigates these issues by explicitly separating perception from reasoning and jointly optimizing both. By generating accurate self-captions, the model grounds its reasoning in the image, leading to more reliable and correct answers.}
\label{hallucinations}
\end{figure}